\documentclass[10pt,twocolumn,letterpaper]{article}

\usepackage[pagenumbers]{wacv} 

\usepackage{graphicx}
\usepackage{amsmath}
\usepackage{amssymb}
\usepackage{booktabs}

\usepackage[utf8]{inputenc} 
\usepackage[T1]{fontenc}    
\usepackage{url}            
\usepackage{amsfonts}       
\usepackage{nicefrac}       
\usepackage{microtype}      
\usepackage{xcolor}         
\usepackage{adjustbox}
\usepackage{subcaption}
\usepackage{enumitem}
\usepackage{caption}

\usepackage[pagebackref,breaklinks,colorlinks]{hyperref}

\usepackage[capitalize]{cleveref}
\crefname{section}{Sec.}{Secs.}
\Crefname{section}{Section}{Sections}
\Crefname{table}{Table}{Tables}
\crefname{table}{Tab.}{Tabs.}

\begin{document}

\title{ComplexVAD: Detecting Interaction Anomalies in Video}

\author{Furkan Mumcu\thanks{Furkan Mumcu did part of this work as an intern at MERL.}\\
University of South Florida\\
{\tt\small furkan@usf.edu}
\and
Michael J. Jones\\
Mitsubishi Electric Research Labs (MERL)\\
{\tt\small mjones@merl.com}
\and
Yasin Yilmaz\\
Univ. of South Florida\\
{\tt\small yasiny@usf.edu}
\and
Anoop Cherian\\
Mitsubishi Electric Research Labs (MERL)\\
{\tt\small cherian@merl.com}
}
\maketitle



\begin{abstract}

Existing video anomaly detection datasets are inadequate for representing complex anomalies that occur due to the interactions between objects. The absence of complex anomalies in previous video anomaly detection datasets affects research by shifting the focus onto simple anomalies. To address this problem, we introduce a new large-scale dataset: ComplexVAD.  In addition, we propose a novel method to detect complex anomalies via modeling the interactions between objects using a scene graph with spatio-temporal attributes. With our proposed method and two other state-of-the-art video anomaly detection methods, we obtain baseline scores on ComplexVAD and demonstrate that our new method outperforms existing works.

\end{abstract}
\section{Introduction}

Video anomaly detection (VAD) has become a popular research area with important security and public safety applications due to the massive amount of video surveillance data being generated which humans cannot effectively monitor.  Video anomaly detection algorithms are crucial for flagging unusual activity in surveillance video for further review by human operators. 
Various formulations of the video anomaly detection problem have been studied by the research community.  In this paper, we focus on the formulation in which nominal videos (also called training videos) containing only normal activities in a particular scene are provided for learning a model.  The goal is to temporally and spatially localize anomalous activity occurring in test video of the same scene.

We are focused on single-scene video anomaly detection because it corresponds to a very common use-case: using a camera to monitor activity at a particular location and alert someone when unusual activity occurs.  
In such scenarios, what is normal in one scene may not be normal in another.  For a new scene, the idiosyncrasies of that scene would need to be learned from video of the new scene and not generalized from other scenes.  For example, something as innocuous as walking across the grass may be anomalous in one scene, but perfectly normal in another.  One can only know this by viewing normal video of a particular scene.  Another important difference between single-scene and multi-scene VAD is the presence of location-specific anomalies in single-scene VAD (i.e. activity that is anomalous in some locations but not others in a scene).  Because there is no overlap in locations for the difference scenes in multi-scene VAD, such datasets do not include location-specific anomalies.  Thus, multi-scene VAD is not a generalization of single-scene VAD.  This is an important point that is often overlooked by researchers in this field.

There have been many datasets introduced for the single-scene version of video anomaly detection, including UCSD Ped1 and Ped2~\cite{mahadevan2010anomaly}, CUHK Avenue~\cite{lu2013abnormal}, Street Scene~\cite{ramachandra2020street}, NOLA~\cite{doshi2022rethinking}, and IITB-Corridor~\cite{RodriguesEtAl2020}.  All of these datasets contain anomalous activity that mainly involves a single object or actor, such as a golf cart driving on a pedestrian walkway, a jaywalker, or a person running, etc.  In the real world, anomalous activity is often not this simple.  In this paper, we introduce the idea of complex anomalies which are anomalies that involve the interaction of two or more objects/actors.  Some examples of complex anomalies include a cyclist running into a car, a person falling off of a skateboard, and a person sitting on a car.  Because existing datasets have very few complex anomalies, we introduce a new dataset, called ComplexVAD, with many different types of anomalies involving interactions between two objects.  By introducing this new dataset, we hope to encourage more complex models of scenes that include modeling of object interactions.  We expect such models to expand the types of anomalies that can be reliably detected. 

In addition to introducing a new dataset and a new direction for video anomaly detection research, we also propose a novel method for detecting complex anomalies in video.
In our method, we generate scene graphs by turning frames into graph representations.  Each object of each frame is extracted (using a multi-class object detector) and treated as a node in the graph where node features are represented by the current location, bounding box, motion trajectory for the next $T$ frames, object class identifier, and skeletal pose if the object is a person. Each node is then connected with nearby nodes in the frame if the 3D spatial distance between objects is below a threshold. 

At the end of this process, we have a graph representation for each frame. We group node-to-node connections which we simply call node pairs or just pairs, into a set of normal pairs. We also collect isolated nodes into another set to detect simple anomalies. Then, we reduce both sets to smaller sets which we call exemplars by removing redundant instances. The details of exemplar selection are given in Section \ref{sec:method}. For a given test video, we again compute scene graphs for test frames and compare node pairs and isolated nodes to the appropriate exemplar set using distance functions between object attributes which are explained in Section \ref{sec:method}.  Any test instance with a high distance to every nominal exemplar is considered anomalous. 

On the ComplexVAD dataset, we compare our proposed method against two state-of-the-art video anomaly detection methods using the frame-level criterion \cite{li2013anomaly}, the region-based detection criterion \cite{ramachandra2020street} and the track-based detection criterion \cite{ramachandra2020street}. Our experimental results show that while our method performs better than existing methods, complex anomaly detection is a difficult problem in needs of further investigation.  

In summary, we make the following key contributions: 
\begin{itemize}
\item We introduce a new large-scale video anomaly detection dataset, named ComplexVAD, to encourage further research on detecting more difficult complex anomalies.
\item We propose a novel video anomaly detection method based on scene graphs to detect complex anomalies.  
\item We demonstrate improved results for our proposed method over two state-of-the-art video anomaly detection methods which establishes a baseline for the new ComplexVAD dataset.
\end{itemize}
\section{Related Work}

There have been many datasets introduced for the problem of video anomaly detection.   UCSD Ped1 and Ped2~\cite{mahadevan2010anomaly} are early datasets with simple anomalies such as cyclists, golf carts and people walking in unusual places.  CUHK Avenue \cite{lu2013abnormal} is another popular dataset whose anomalies include people running or walking in unusual directions or throwing things into the air.  Street Scene \cite{ramachandra2020street} emphasizes location-dependent anomalies such as people jaywalking, cars parked illegally or cars/bikes moving outside of their designated lanes.  IITB-Corridor \cite{RodriguesEtAl2020} is a dataset with anomalies such as loitering, left-behind luggage, people running and people fighting. NOLA \cite{doshi2022rethinking} is another dataset proposed to study continual learning in VAD.  These datasets are all single-scene datasets which is our main focus in this paper.

There are also a number of datasets intended for anomaly detection across multiple scenes. 
ShanghaiTech \cite{luo2017revisit} includes videos of 13 different scenes with anomalies such as cyclists, people with strollers, and people fighting.
UCF-Crime \cite{sultani2018real} is another multi-scene dataset intended for weakly supervised version of video anomaly detection in which anomalous videos are used in addition to normal videos during training.  Anomaly types include people fighting and explosions.  UBnormal \cite{ubnormal} is a multi-scene dataset consisting of synthetically generated scenes that include annotated anomalies in the training videos.  Anomalous activity includes people running, falling, dancing and jaywalking.

The vast majority of anomalies in all of these datasets (with the exception of UCF-Crime) involve a single object, for example, a person walking in an unusual place, the appearance of a golf cart, a cyclist, or a person running.  Such anomalies can be detected well (at least for temporal localization) by models that fundamentally work at the pixel level as evidenced by so many models that use pixel reconstruction error as a loss function for training \cite{hasan2016learning,ionescu2019object,nguyen2019anomaly,chang2020clustering,lu2020few,liu2021hybrid,liu2018future,wang2021prediction,yang2023video,ShiEtAl2023,yangetal2023,SunGong2023}.  Concerning the UCF-Crime dataset \cite{sultani2018real}, it is designed for a very different version of video anomaly detection (multi-scene and weakly supervised) which does not correspond to the most common real-world surveillance application that we are most interested in.  We hope to encourage methods that try to understand a scene at a higher level such as methods that model objects and their motions.  Toward that end, a dataset that has more complex anomalies such as those involving the interaction of multiple objects will require modeling a scene at a higher level to be successful.  This is the main motivation for introducing our new ComplexVAD dataset.

The novel algorithm we propose for detecting complex anomalies uses a scene graph to represent objects and their interactions in a video.   A number of recent papers have also focused on addressing anomaly localization at the object-level \cite{georgescu2021anomaly, georgescu2021background, ionescu2019object, barbalau2023ssmtl++, yu2020cloze, wang2022video, doshi2020any, doshi2021efficient, doshi2022modular}. These methods utilize pre-trained object detectors to first localize objects and then estimate if the detected objects are anomalous or not.  There are many differences in the details of these methods compared to ours, especially in the representation of motion, but the most important difference is that these methods do not model the interactions among people/objects. 

There have been a few past approaches that did model interactions among objects.  Many of these methods also employed scene graphs to represent the interactions \cite{ChenEtAl2018,doshi2023interpretable,SunEtAl2020}.  In \cite{ChenEtAl2018}, a simple model of object-relation-object triplets is used to model a scene, but unlike our method, there is no modeling of motion or trajectories.  In our approach, the trajectory for each object is computed which allows unusual trajectories to be detected as anomalous.  The approach of \cite{doshi2023interpretable} used a scene graph to represent subject-predicate-object triplets in normal video and then compare those to ones found in test video.  The main difference compared with our proposed approach is our use of an exemplar-based model of normal pairs of objects and our inclusion of object trajectories in the representation of objects in the scene graph.  In \cite{SunEtAl2020}, a scene graph was also used to represent objects and their interactions.  The main difference with our approach is the method for computing distances between pairs of graph nodes and the specific attributes that are stored in the representation of each object.

The work of \cite{SzymanowiczEtAl2021} modeled interactions between a person and an object using human-object interaction (HOI) vectors that does not use scene graphs.  Normal HOI vectors are modeled with a Gaussian mixture.  Low probability HOI vectors from test video can then be detected as anomalous.

Most of the methods that use object-level representations including our method are also interpretable.  They can provide human-understandable explanations for detected anomalies.  Explainability is a very important property for VAD methods to be adopted for real-world use.

\section{ComplexVAD}

\begin{table*}[t!]

  \centering
  \begin{adjustbox}{width=0.8\linewidth,center}
  \begin{tabular}{|c | c | c | c  |c | c | c | c | c |}
    \hline
    Dataset & \begin{tabular}{@{}c} Total\\Frames \end{tabular}  & \begin{tabular}{@{}c} Training\\Frames \end{tabular}  & \begin{tabular}{@{}c} Testing\\Frames \end{tabular} & \begin{tabular}{@{}c} Anomalous\\Events \end{tabular} & \begin{tabular}{@{}c} Anomaly\\Types \end{tabular} & \begin{tabular}{@{}c} Ground\\Truth \end{tabular} & Resolution & \begin{tabular}{@{}c} Complex\\Anomalies \end{tabular} \\
    \hline
    UCSD Ped1 & 14,000 & 6800 & 7200 & 54 & 5 & S, T & 238 x 158 & No\\
    \hline
    UCSD Ped2 &  	4560 & 2550 & 2010 & 23  &5 & S, T & 360x240 & No\\ 
    \hline
    CUHK Avenue  & 30,652 & 15,328 & 15,324 & 47 & 5 & S, T & 640 x 360 & No \\
    \hline
    IITB-Corridor &  483,566 &301,999 & 181,567  & ? & $\sim$10 & T& 1920x1080 & No \\
    \hline
    NOLA  &  	1,440,000 & 450,000 & 990,000 & 50 & $\sim$10 & S, T & 1280 x 720 & No \\
    \hline
    Street Scene & 203,257 & 56,847 & 146,410 & 205 & 17 & S, T & 1280 x 720 & No \\ 
    \hline
    \textbf{ComplexVAD} &  \textbf{3,681,438}	&  \textbf{2,069,941} &   \textbf{1,611,497} & \textbf{118}	& \textbf{40}  & \textbf{S, T} & \textbf{1920x1080} & \textbf{Yes}\\
    \hline
    
  \end{tabular}
  \end{adjustbox}
  \caption{Characteristics of existing single scene video anomaly detection datasets compared to ComplexVAD. S and T denote Spatial and Temporal ground truth labels respectively. }
  \label{tab:comparison}
  \vspace{-7mm}
\end{table*}

\begin{figure}[!tb]
\centering
\minipage{0.25\textwidth}
  \includegraphics[width=\linewidth]{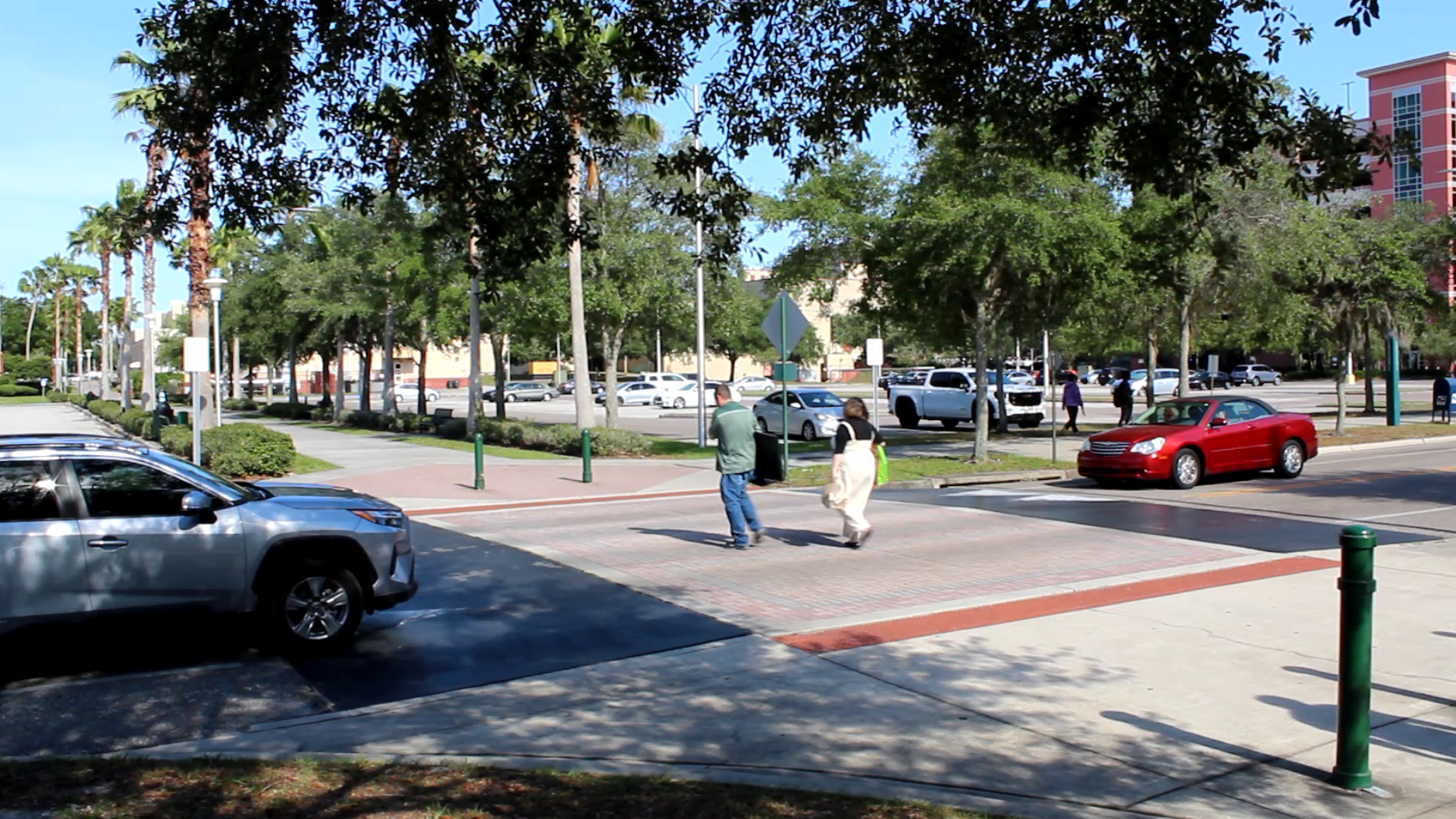}
\endminipage\hfill

\bigskip

\minipage{0.25\textwidth}
  \includegraphics[width=\linewidth]{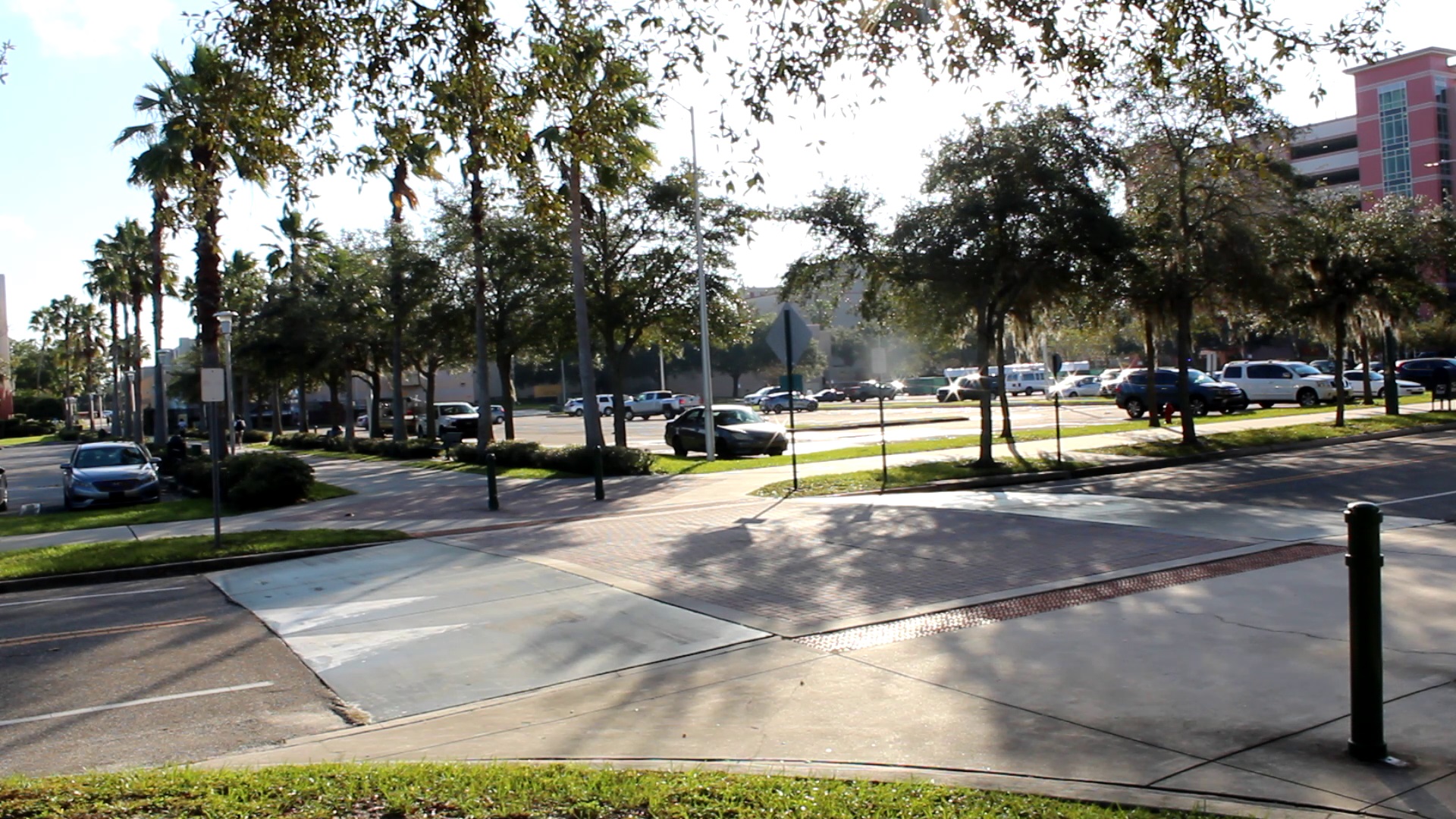}
\endminipage\hfill

\bigskip

\minipage{0.25\textwidth}%
  \includegraphics[width=\linewidth]{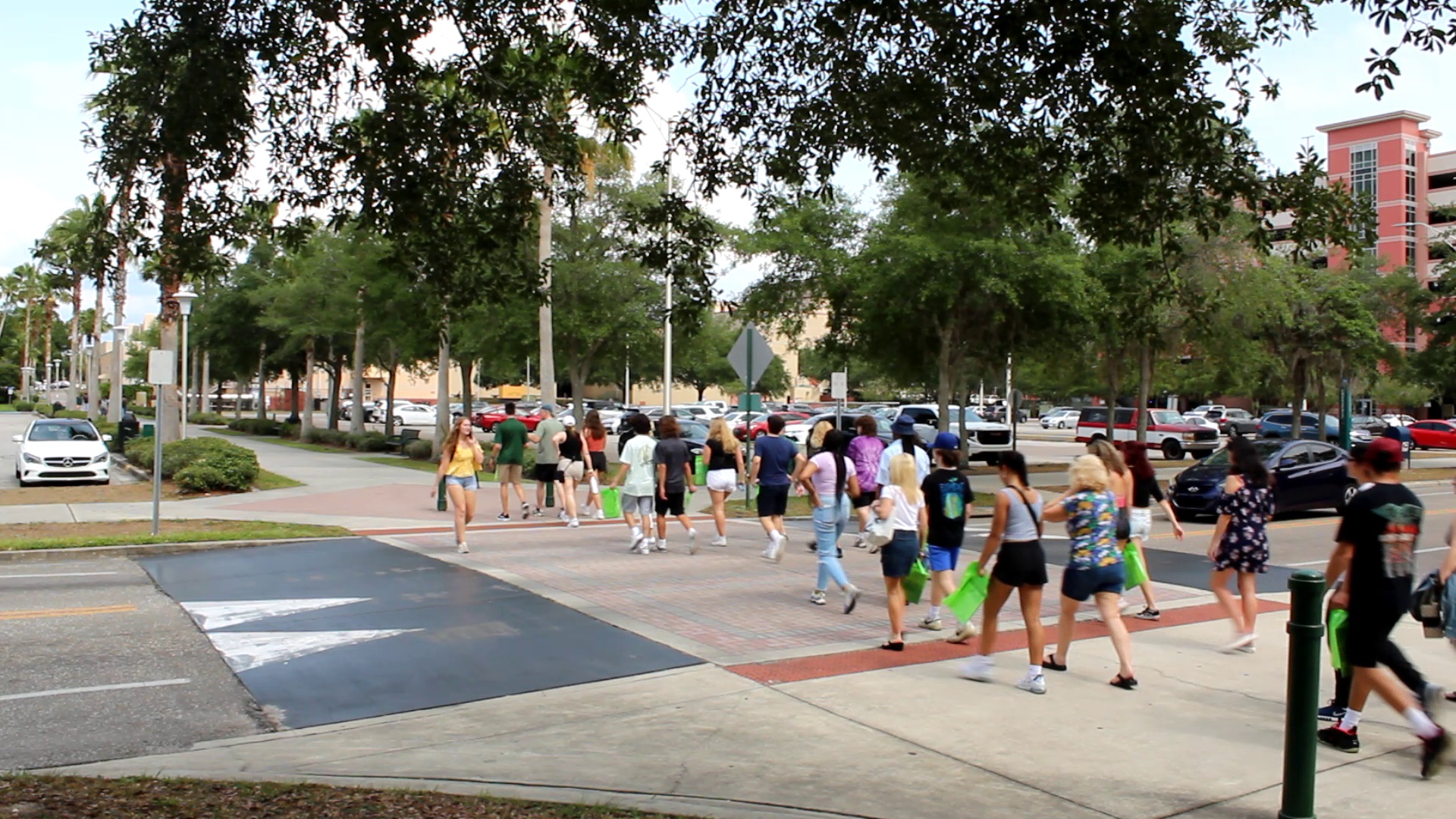}
\endminipage
\caption{Nominal frame samples from the ComplexVAD dataset.}
\label{fig:nominal}
\vspace{-7mm}
\end{figure}

\begin{figure}[tb]
\centering
\minipage{0.25\textwidth}
  \includegraphics[width=\linewidth]{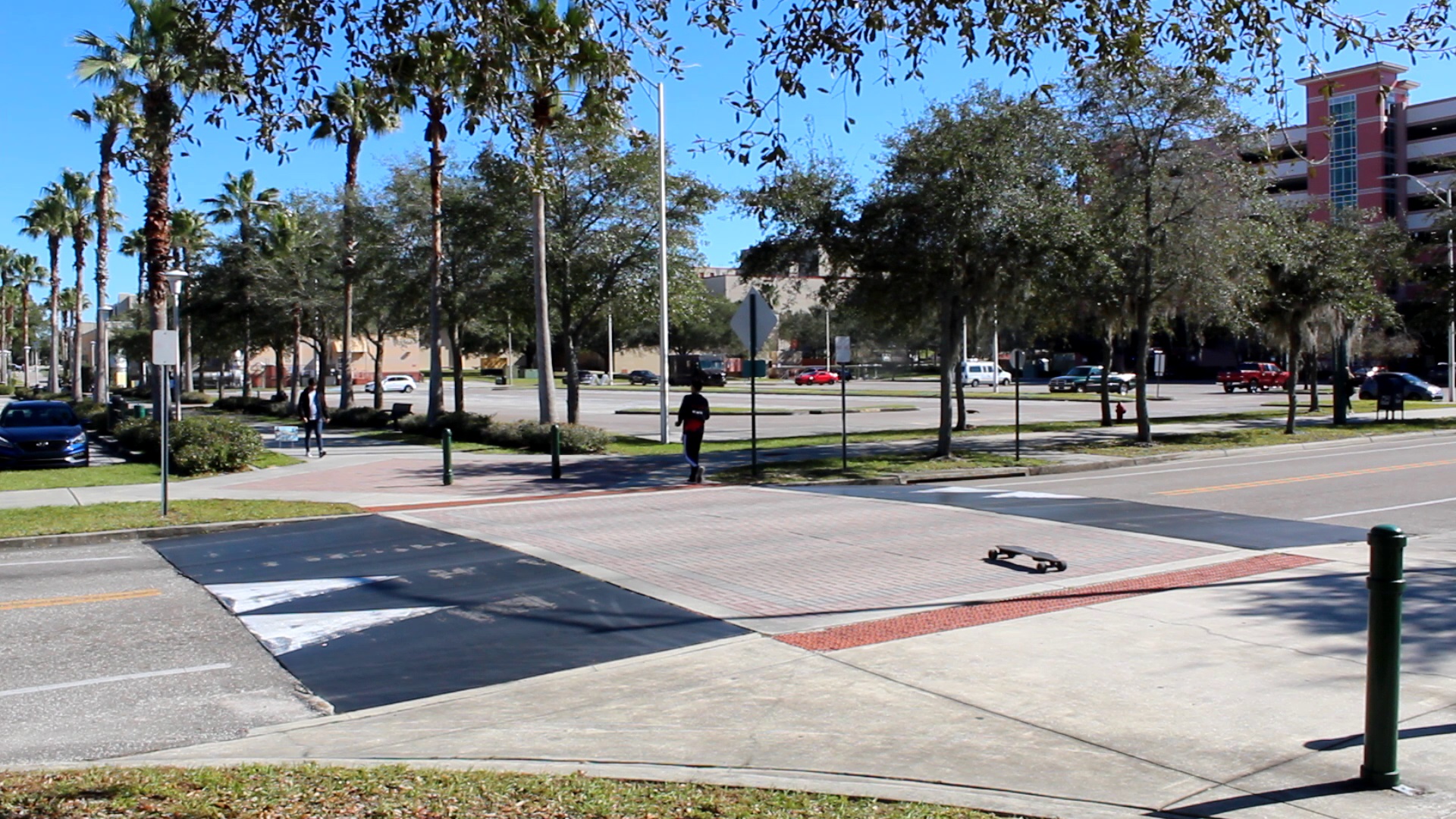}
\endminipage\hfill

\bigskip

\minipage{0.25\textwidth}
  \includegraphics[width=\linewidth]{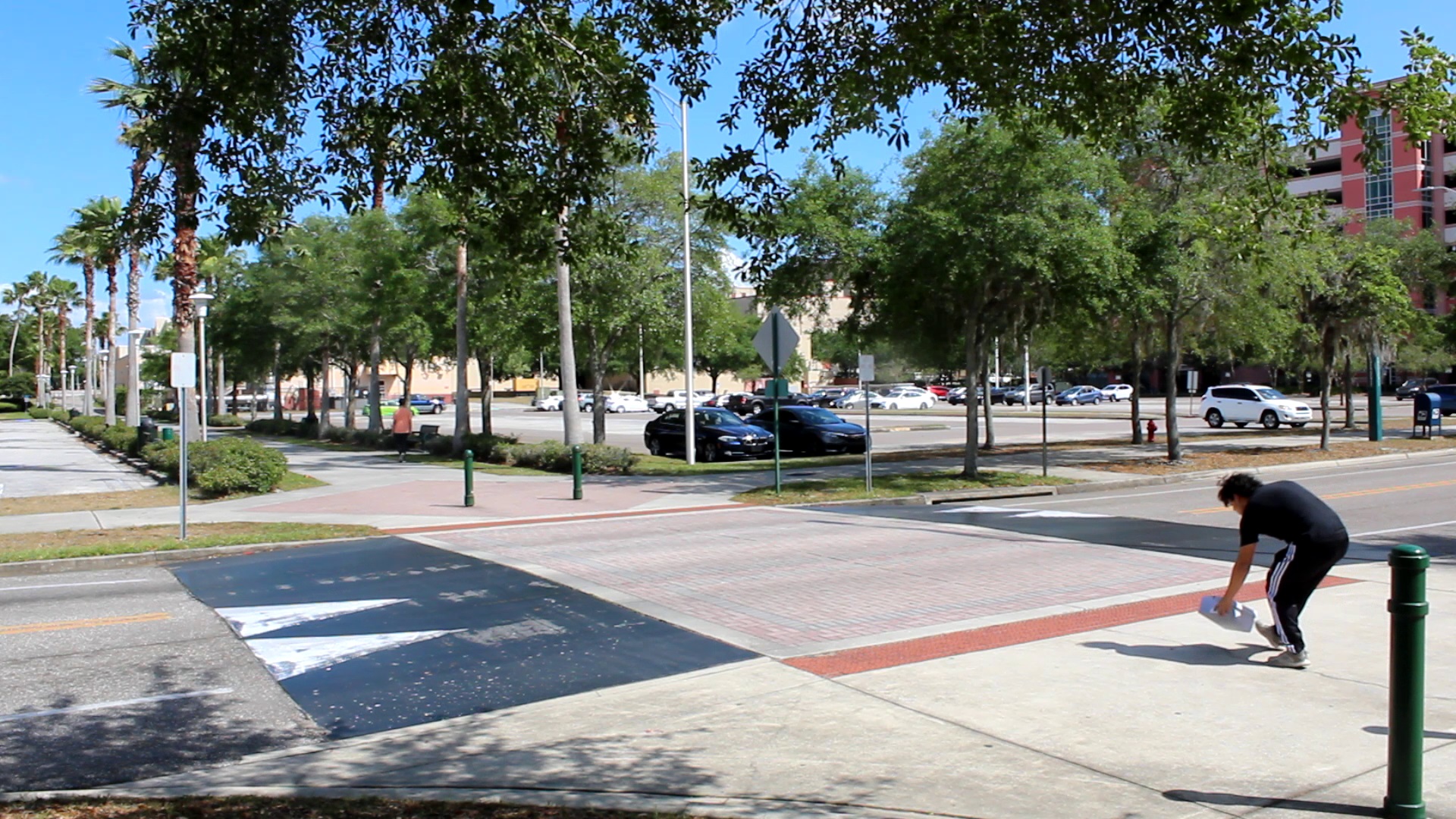}
\endminipage\hfill

\bigskip

\minipage{0.25\textwidth}%
  \includegraphics[width=\linewidth]{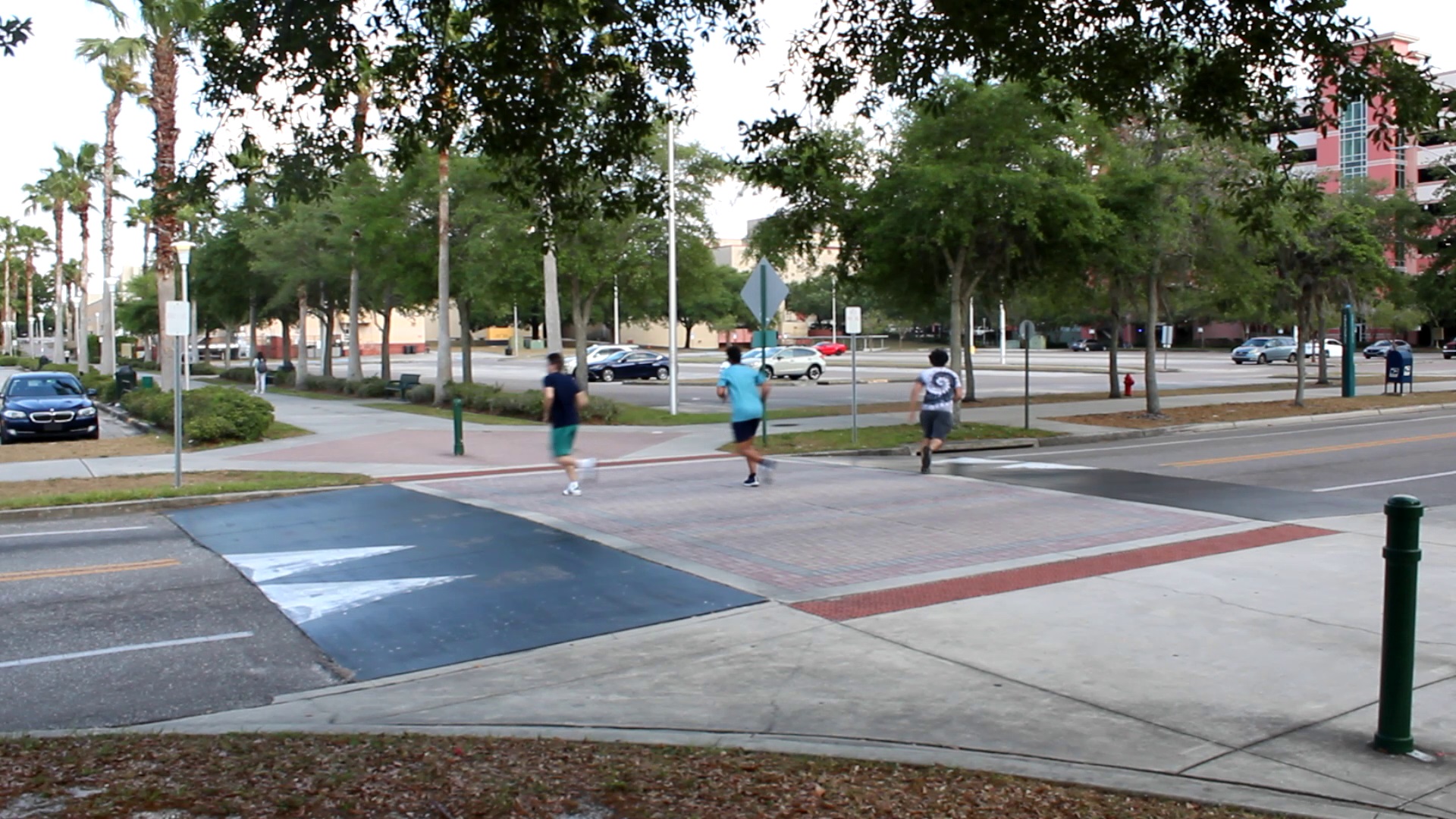}
\endminipage
\caption{Samples of complex anomaly from the ComplexVAD dataset. In all samples, objects are nominal, but the interactions are anomalous. (Top) A skateboard moving alone violates the expectation of a nominal interaction between a person and skateboard. (Middle) Person carries an object and then leaves it on the ground. (Bottom) Three people walk together and then suddenly start to run in different directions.}
\label{fig:anomalous}
\vspace{-7mm}
\end{figure}

To address the absence of complex anomalies in existing datasets, we introduce the ComplexVAD dataset. The dataset has 104 training and 113 test video sequences. All videos are recorded at the same location in a university campus showing a crosswalk, pedestrian sidewalks, and a two-lane street. Figure \ref{fig:nominal} shows some nominal frames. The video collection process lasted 5 months and videos were recorded during different periods including morning, noon, and afternoon. Since it is a campus environment, the scene tends to change frequently depending on the time and day. For this reason, for each day of the week, there is at least one hour of recording for morning, noon, and afternoon to represent the different states of the scene. It is a highly active and complex scene with people who walk, jog, or run; bikers, skateboarders, and scooter riders using the crosswalk and sidewalks; cars, buses, and golf carts using the car lanes. In addition, the background is not static among videos due to changing shadows, trees blowing in the wind or parking lots with varying numbers of parked cars. All faces were blurred using a face detector and Gaussian blurring to remove personally identifiable information.

ComplexVAD is a large dataset consisting of videos recorded in 1080x1920 resolution and at a rate of 30 frames per second. The training set includes videos ranging from 2.5 minutes to 13 minutes, with an average duration of 11 minutes. In the test set, the longest duration is 12.8 minutes, the shortest is 1.5 minutes, and the average duration is 7.9 minutes. When considering frames extracted from the original videos at 30 frames per second, there are 2,069,941 RGB frames for training and 1,611,497 RGB frames for testing, totaling 3,681,438 RGB video frames for the entire dataset. Due to potential complications and challenges in storage and distribution, ComplexVAD is publicly shared in video format.  The comparison with existing VAD datasets can be seen in Table \ref{tab:comparison}. 

The aim of the ComplexVAD dataset is to showcase complex anomalies. We define a complex anomaly as an anomalous event resulting from the interaction between objects. Compared to anomalies presented in previous datasets, objects in a complex anomaly should be considered normal within the scene until their interaction occurs. Some examples of complex anomaly are presented in Figure \ref{fig:anomalous}. For instance, a person and a backpack are common objects in our dataset, but a person leaving their backpack on a sidewalk constitutes an anomaly resulting from the "leaving" action. Another example is a skateboard moving autonomously (due to a remote control), on a crosswalk. While skateboards are typically found in crosswalks with someone riding them, in this case, the usual interaction between the skateboard and a rider is absent.

Additionally, changes in interactions can lead to complex anomalies, such as a biker slowing down and stopping briefly in the middle of a crosswalk, where the typical interaction involves passing by without any interruption. The ComplexVAD dataset includes complex anomalies resulting from interactions between various objects such as pedestrians, cars, bikes, scooters, skates, sports balls, dogs, baseball bats, and trees. ComplexVAD includes 118 anomalies from 40 diverse types of complex anomalies, which are detailed in the supplementary document.

The ComplexVAD dataset is publicly available under the CC-BY-SA-4.0 license.\footnote{\url{www.merl.com/research/downloads/ComplexVAD}} We provide ground truth annotations in a form which can easily be used for several types of evaluation criteria such as region-based and track-based as well as frame-level. Annotations are provided for each testing video in the form of bounding boxes around each object that is a part of the anomalous event in each frame. In addition, a track id is assigned to each bounding box so that each anomalous event can be represented as a track of bounding boxes. Due to the nature of our dataset, each frame can have more than one anomaly labeled.
\section{Detecting Complex Anomalies }
\label{sec:method}

\begin{figure*}[t]
  \centering
   \includegraphics[scale=0.44]{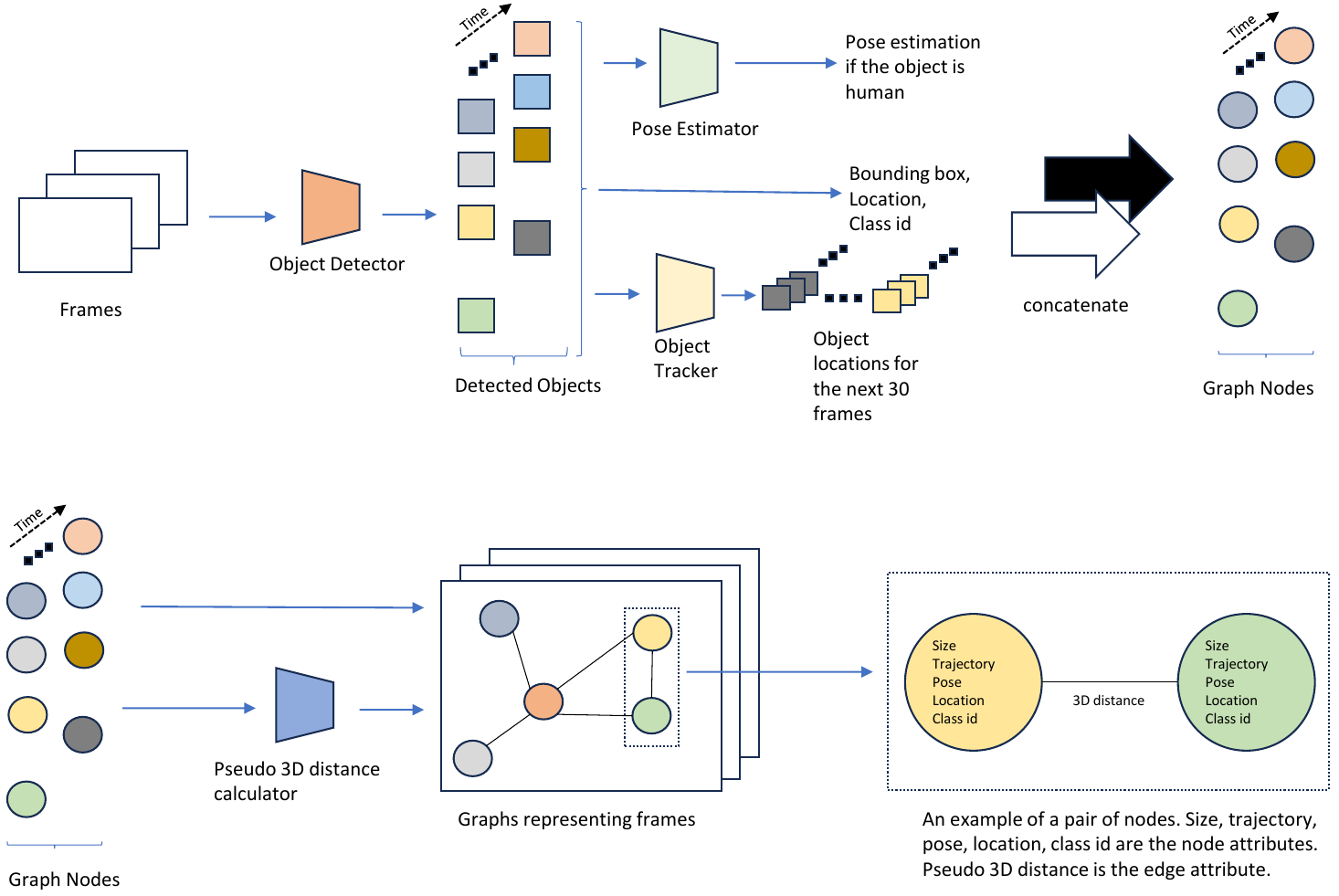}
   \caption{The pipeline of our method for frame to graph generation with the help of an object detector, object tracker and pose estimator.}
   \label{fig:frame_to_graph}
   \vspace{-5mm}
\end{figure*}

We propose a novel method to detect complex anomalies. Our method can be divided into three stages. 
First, we derive graph representations of all frames in the training dataset. Second, for all pairs of nodes (i.e., pairs of objects that are close in terms of 3D distance) and isolated nodes (i.e., objects that are not close to any other object) in the training set, we use an exemplar selection algorithm to select a subset of unique node pairs and isolated nodes to form an exemplar set.  Third, we compare the distances between node pairs in the test set and node pairs in the exemplar set.  The same is done between isolated test nodes and isolated nodes in the exemplar set.  Any test instance with a high distance to every exemplar is considered anomalous. In the following sections, we will discuss the stages of our method in detail. 

\subsection{Frame to graph}
\label{frame_to_graph}

For a given dataset, we transform each frame of each video into an undirected graph. The pipeline of our frame to graph transformation is depicted in Figure~\ref{fig:frame_to_graph}. Our first step is to use an object detector to extract objects. Note that, for our approach, the object detector plays a fundamental role. It is important to evaluate the object detector's capability in the scene and choose the most suitable one. In our initial experiments, we found that Detectron2 \cite{wu2019detectron2} has the most accurate object detection. Hence, Detectron2, which is trained on the COCO dataset, is used in our implementation.

A video $V$ is a collection of $M$ frames $\{F_i\}_{i=1}^M$, such that $V = [F_1, F_2,..., F_M]$. We send each frame $F_i$ to the object detector $O$, which returns $X$ number of detected objects. For each object $o$, the location $l = (x_o, y_o)$ which is the $x$ and $y$ coordinates of the center of the object, $b = (w_o, h_o)$ which is the width and height of the bounding box for the object, and class id $c$. The output of the object detector is then $O(F) = [o_1, o_2,..., o_X]$, where each object $o_i$ is represented by $o_i = [b, c, l]$.

After detecting objects in a frame, they are then tracked using an object tracker, namely ByteTrack \cite{zhang2022bytetrack}. Each detected object $o$ is sent to the object tracker, which returns $x$ and $y$ coordinates for that object in the subsequent frames. In our method, we track objects for 30 frames. Therefore, for every object, we acquire the trajectory $\theta = \{(x_1, y_1), (x_2, y_2) ,...,(x_{30}, y_{30})\}$.

In addition to the object detector and object tracker, we also use a pose estimator to obtain the pose information of human objects. Any object $o$ identified as human is sent to the pose estimator to obtain the pose vector $p = \{(x_1, y_1), (x_2, y_2) ,...,(x_{17}, y_{17})\}$, which contains the locations of 17 key points on the human body.  In our experiments, we use the human pose estimation method included in Detectron2 \cite{wu2019detectron2}.

We concatenate the features that we extract from an object $o$ to form a graph node $n = [b, c, l, \theta, p]$ where node attribute $b$ is the bounding box size, $c$ is the class id, $l$ is the location of the center of the object, $\theta$ is the trajectory vector and $p$ is the pose vector.

Next, we need to find edges between nodes/objects that are likely to be interacting.  Past approaches to building scene graphs \cite{ChenEtAl2018,SunEtAl2020} have used a deep network, usually trained on the Visual Genome dataset \cite{visualgenome}, to estimate relations between objects.  We found such approaches to produce too many inconsistent relations which can cause false positive anomalous detections. 
Instead we use the simple and more robust method of assigning an edge between objects if they are close to each other (if their distance in 3D space is below a threshold).  Thus, to determine which nodes to connect in the graph we need to calculate the 3D distances between each pair of nodes. To calculate the 3D distance we need to derive the 3D coordinates of the node locations by estimating a pseudo-depth since we do not have access to actual depth estimates. Given two nodes $n_1$ and $n_2$, we have 2D coordinates $l_1 = (x_1, y_1)$ and $l_2 = (x_2, y_2)$. Then we define a relative depth, $z$, between two nodes by taking the absolute difference of $y$ values such that $z = |y_1 - y_2|$. This estimate of pseudo-depth assumes that objects are resting on the ground plane and the ground plane is farther from the camera the closer it is to the top of the image. The 3D distance $d$ can then be calculated by taking the Euclidean distance between 3D coordinates $(x_1, y_1, z)$ and $(x_2, y_2, 0)$. Any node pair that has a 3D distance $d$ smaller than a predetermined threshold $h$ is connected with an edge $E$. Due to applying the threshold, not every single node is necessarily connected to another node, which leads to having isolated nodes in addition to node pairs.

At the end of our frame to graph transformation, a frame $F$ which is a collection of objects $\{o_1, o_2,..., o_X\}$ where $X$ is the total number of objects extracted by the object detector, can be represented as a graph $G = (N, E)$ where $N$ is the collection of graph nodes $[n_1, n_2,..., n_X]$, and $E$ is the graph edges between connected nodes. Similarly, a video $V = [F_1, F_2,..., F_M]$ which contains $M$ number of frames $F$, can be represented as collection of graphs: $V = [G_1, G_2,..., G_M]$.

\subsection{Model building from nominal video}

For a given nominal video, frames are processed using our method described in \ref{frame_to_graph} and transformed into graphs. For all frames in a video, we collect all pairs of nodes that are connected by an edge into one set and all isolated nodes (not connected to any other node) into another set. Then for each of the sets, independently, we run an exemplar selection algorithm which selects a subset of the elements of the set such that no two members of the subset are near each other according to a distance function (described below).  The intuition behind exemplar selection is to simply remove redundant (or nearly redundant) elements from the set leaving behind a compact, representative subset of exemplars.  We use the same exemplar selection algorithm as described in \cite{ramachandra2020street}. Given a set $S$, the exemplar selection algorithm proceeds as follows: (1) Initialize the exemplar set to NULL. (2) Add the first element of $S$ to the exemplar set. (3)  For each subsequent element of $S$, find its distance to the nearest instance in the exemplar set. If this distance is above a threshold, $th$, then add the element to the exemplar set.

As mentioned before, we run exemplar selection separately on the set of all isolated nodes found in the graphs of all frames and the set of all pairs of nodes found in the graphs of all frames.  To use the exemplar selection algorithm we need to define a distance between two isolated nodes and a distance between two node pairs.  We will start with the distance between two isolated nodes.

A graph node, $n$,  is a high-level representation of an object which includes the attributes $[b, c, l, \theta, p ]$ where $b$ is the bounding box size, $c$ is the class identifier, $l$ is the location, $T$ is the trajectory vector and $P$ is the pose vector. For two given nodes $n_1$ and $n_2$ with attributes $[b_1, c_1, l_1, \theta_1, p_1 ]$ and $[b_2, c_2, l_2, \theta_2, p_2]$, we define a distance between each node attribute as follows.

The location distance is the Euclidean distance between $l_1 = (x_1, y_1)$ and $l_2 = (x_2, y_2)$:

\begin{equation}  
\label{eq:location}
L(n_1, n_2) = \sqrt{(x_1 - x_2)^2 + (y_1 - y_2)^2}
\end{equation}

The distance between bounding box sizes $b_1 = (w_1, h_1)$ and $b_2 = (w_2, h_2)$ is calculated by taking the Euclidean distance between each bounding box width and height normalized by the minimum width and height:

\begin{equation}  
S\left(n_1, n_2\right)=\sqrt{\frac{\left(w_1-w_2\right)^2}{\min \left(w_1, w_2\right)}+\frac{\left(h_1-h_2\right)^2}{\min \left(h_1, h_2\right)}}
\end{equation}

The class distance is set to $0$ if the nodes have the same class id; otherwise, it is set to $1$:

\begin{equation}
C(n_1, n_2) = 
\begin{cases}
    0 & \text{if $c_1 = c_2 $} \\
    1 & \text{if $c_1 \neq c_2 $}, \\
\end{cases}
\end{equation}

For two pose vectors, $P_1 = \{(x_{1,1}, y_{1,2}), (x_{1,2}, y_{1,2}) ,...,(x_{1,17}, y_{1,17})\}$ and\\
$P_2 = \{(x_{2,1}, y_{2,2}),(x_{2,2}, y_{2,2}) ,...,(x_{2,17}, y_{2,17})\}$, The pose distance is

\begin{equation}
    P\left(n_1, n_2\right)=\sum_{t=2}^{17} \frac{\left|d p_{1, t}-d p_{2, t}\right|}{\max \left(\min \left(d p_{1, t}, d p_{2, t}\right), 1\right)}
    \label{eq:pose}
\end{equation}
where 
\begin{equation}
d p_{1,t}= \sqrt{(x_{1,1} - x_{1,t})^2 + (y_{1,1} - y_{1,t})^2}
\end{equation}
and
\begin{equation}
d p_{2,t}= \sqrt{(x_{2,1} - x_{2,t})^2 + (y_{2,1} - y_{2,t})^2}
\end{equation}
are the distances from the first pose keypoint to the $t$th pose keypoint, for each pose vector, respectively.  The $max$ function in the denominator of Equation \ref{eq:pose} insures that the denominator is not less than 1 to prevent division by zero.

For two node trajectories $\theta_1 = \{(x_{1,1}, y_{1,1}), (x_{1,2}, y_{1,2}) ,...,(x_{1, 30}, y_{1, 30})\}$ and\\
$\theta_2 = \{(x_{2,1}, y_{2,1}), (x_{2,2}, y_{2,2}) ,...,(x_{2, 30}, y_{2, 30})\}$, the trajectory distance is the sum of the L1 distances between the displacements of the first node and the displacements of the second node normalized by the minimum displacement:

\begin{multline}
\Theta\left(\theta_1, \theta_2\right)=\sum_{t=1}^{T-1} \frac{\left|d x_{1, t}-d x_{2, t}\right|}{\max \left(\min \left(d x_{1, t}, d x_{2, t}\right), 1\right)} + \\
\frac{\left|d y_{1, t}-d y_{2, t}\right|}{\max \left(\min \left(d y_{1, t}, d y_{2, t}\right), 1\right)}
\end{multline}

where $dx_1(t) = x_{1,t}-x_{1,t+1}$, $dx_2(t) = x_{2,t}-x_{2,t+1}$, $dy_1(t) = y_{1,t}-y_{1,t+1}$, $dy_2(t) = y_{2,t}-y_{2,t+1}$.  $T$ is the number of frames in a track which is set to 30 in our experiments.  The max function in the denominator is used to avoid division by zero.

Given these distances between attributes of two nodes, the final distance between two isolated nodes is calculated as follows:

\begin{multline}
D(n_1, n_2) = \max( \frac{L(n_1, n_2) - \mu_L}{\sigma_L},  \\
\frac{S(n_1, n_2) - \mu_S}{\sigma_S}, \frac{C(n_1, n_2) - \mu_C}{\sigma_C}, \\
\frac{P(n_1, n_2) - \mu_P}{\sigma_P}, \frac{\Theta(n_1, n_2) - \mu_\Theta}{\sigma_\Theta} )
\label{eq:dist}
\end{multline}
where the $\mu$ and $\sigma$ parameters are normalization constants for each distance which make all the distances comparable.  We discuss how these normalization constants are chosen in the supplementary material.

A node pair $N$ is a combination of two nodes which are connected with an edge. Between two node pairs $N_1 = (n_1, n_2)$ and $N_2 = (n_3, n_4)$, the distance is calculated as follows:

\begin{multline}
D_{pair}(N_1, N_2) = \\
\min(\max(D(n_1, n_3), D(n_2, n_4)), \\
\max(D(n_1, n_4), D(n_2, n_3))).
\end{multline}

The intuition behind this distance is firstly that we do not know whether $n_1$ corresponds to $n_3$ or $n_4$ (and similarly whether $n_2$ corresponds to $n_3$ or $n_4$) so we need to try both pairings and take the minimum distance.  This corresponds to the outer $min$ function.  For a given correspondence, the overall distance between the two node pairs is the maximum distance between the corresponding nodes from each pair.  This is represented by the inner $max$ functions. Further details on distance normalization and exemplar selection provided in the supplementary document.

\subsection{Complex anomaly detection in test video}

After the first stage of obtaining exemplar sets from nominal videos, the next step is detecting anomalies in testing video of the same scene. As with nominal videos, the pipeline that is described in \ref{frame_to_graph} is also followed for test videos, to generate graphs from objects detected in each frame. The same object attributes are computed for each object: location, bounding box size, class ID, trajectory and if the object is a person, a pose vector. Given a scene graph for a test frame, anomaly scores are computed for every pair of connected nodes and for every isolated node.
The anomaly score, $AS$, for a test isolated node, $n$, is the distance to the nearest exemplar in the isolated node exemplar set:

\begin{equation}
    AS(n, \mathcal{E}_{iso})=\min _{n_e \in \mathcal{E}_{iso}} D\left(n, n_e\right)
    \label{eq:nniso}
\end{equation}

Similarly, the anomaly score (AS) for a pair of nodes $N=(n_1, n_2)$ is the distance to the nearest pair of nodes from the node-pair exemplar set:

\begin{equation}
    AS(N, \mathcal{E}_{pair})=\min _{N_e \in \mathcal{E}_{pair}} D_{pair}\left(N, N_e\right)
    \label{eq:nnpair}
\end{equation}

The nearest neighbor searches in Equations \ref{eq:nniso} and \ref{eq:nnpair} are generally fast because the number of exemplars is typically small, but can easily be sped up with one of the many efficient nearest neighbor techniques \cite{flann}.
\section{Experiments}

\subsection{Experimental settings and evaluation criteria}

We evaluate our proposed method and two other state-of-the-art video anomaly detection methods, namely Memory-augmented Deep Autoencoder (MemAE) \cite{MemAE2019} and Explainable Video Anomaly Localization (EVAL) \cite{SinghEtAl2023}, on the ComplexVAD dataset. 

For our method, ByteTrack \cite{zhang2022bytetrack} and Detectron2 \cite{wu2019detectron2} were used as object tracker and object detector. The pose estimator module of Detectron2 is also used for pose estimations. Using the  method described in Section \ref{sec:method}, exemplar sets are extracted for all of the training videos. We choose a threshold $th = 0.65$ for exemplar selection that resulted in a modest number of total exemplars selected. From past work that used exemplar-based models, this threshold mainly effects model size and has a small effect on test accuracy.

To test MemAE on the ComplexVAD dataset, we used the same proposed hyper-parameter and model structure settings as described in \cite{MemAE2019}. The ComplexVAD dataset is resized to 256x256 to be compatible with the existing settings. Additionally, since the ComplexVAD dataset has an extensive number of frames, we sub-sampled every third frame of the dataset for training and testing to gain computational speed during training and testing. Finally, the MemAE model is trained with the training split on NVIDIA 4090.

To test EVAL on the ComplexVAD dataset, we subsampled every other frame (for an effective frame rate of 15 fps), and used 10 frame video volumes with 256x256 pixel spatial region sizes which roughly corresponds to the average height of a person in this dataset.  The remainder of the setup and parameters were exactly as described in \cite{SinghEtAl2023}.

We use the Region-Based Detection Criterion (RBDC) and the Track-Based Detection Criterion (TBDC) as proposed in \cite{ramachandra2020street} as our primary evaluation criteria and report the area under the curve (AUC) for false positive rates per frame from 0 to 1. We also report frame-level AUC\cite{mahadevan2010anomaly} scores. As highlighted in previous works \cite{ramachandra2020street} frame-level AUC only evaluates temporal accuracy and disregards spatial localization of anomalies. Whereas, RBDC and TBDC measure a method's capacity to accurately identify anomalous spatio-temporal regions within a given video sequence, however, we also report frame-level AUC scores of the methods for completeness, as well as comparisons to older methods.

In order to get RBDC and TBDC numbers for MemAE we used the following procedure.  For each anomaly score threshold, we create a mask of all pixels with anomaly scores above threshold.  We then find connected components of anomalous pixels.  This give us anomalous regions.  For each connected component with at least 10 pixels, we compute the minimum bounding box encompassing that component.  This yields a set of anomalous bounding boxes that can be used for computing RBDC and TBDC numbers.

\subsection{Results}

\begin{table}[t]
  \centering
  \setlength{\tabcolsep}{10pt} 
  \renewcommand{\arraystretch}{1.2} 
  \begin{tabular}{| c | c | c | c |}
    \hline
    Method &  RBDC & TBDC & Frame \\ 
    \hline
    MemAE \cite{MemAE2019} &  0.0005 &  0 &  0.58 \\
    \hline
    EVAL \cite{SinghEtAl2023} &  0.10 & 0.62 & 0.54 \\
    \hline
    Ours &  {\bf 0.19} &  {\bf 0.64} & {\bf 0.60} \\
    \hline
    
  \end{tabular}
  \caption{The table reports the area under the curve (AUC) for our method and two recent VAD methods using the RBDC, TBDC and Frame-Level evaluation criteria on ComplexVAD.}
  \label{tab:comparison-auc}
\vspace{-7mm}
    
\end{table}

\begin{figure*}[t]
  \centering
   \includegraphics[width=\linewidth]{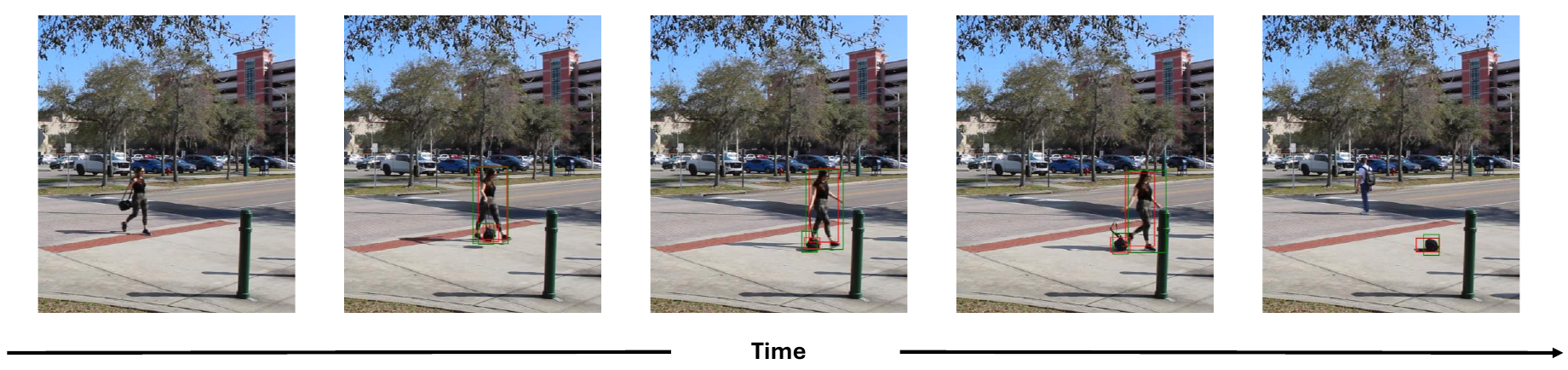}
   \caption{A person who drops a bag on the street is detected as an anomaly with our method. The detection starts with the action of "drop". After the object interaction ends, the dropped object continues to be detected as an anomaly. Ground truth labels and detection boxes are represented with green and red colors, respectively.}
   \label{fig:droppedbag}
   \vspace{-4mm}
\end{figure*}

\begin{figure}[t]
\centering
\minipage{0.25\textwidth}
  \includegraphics[width=\linewidth]{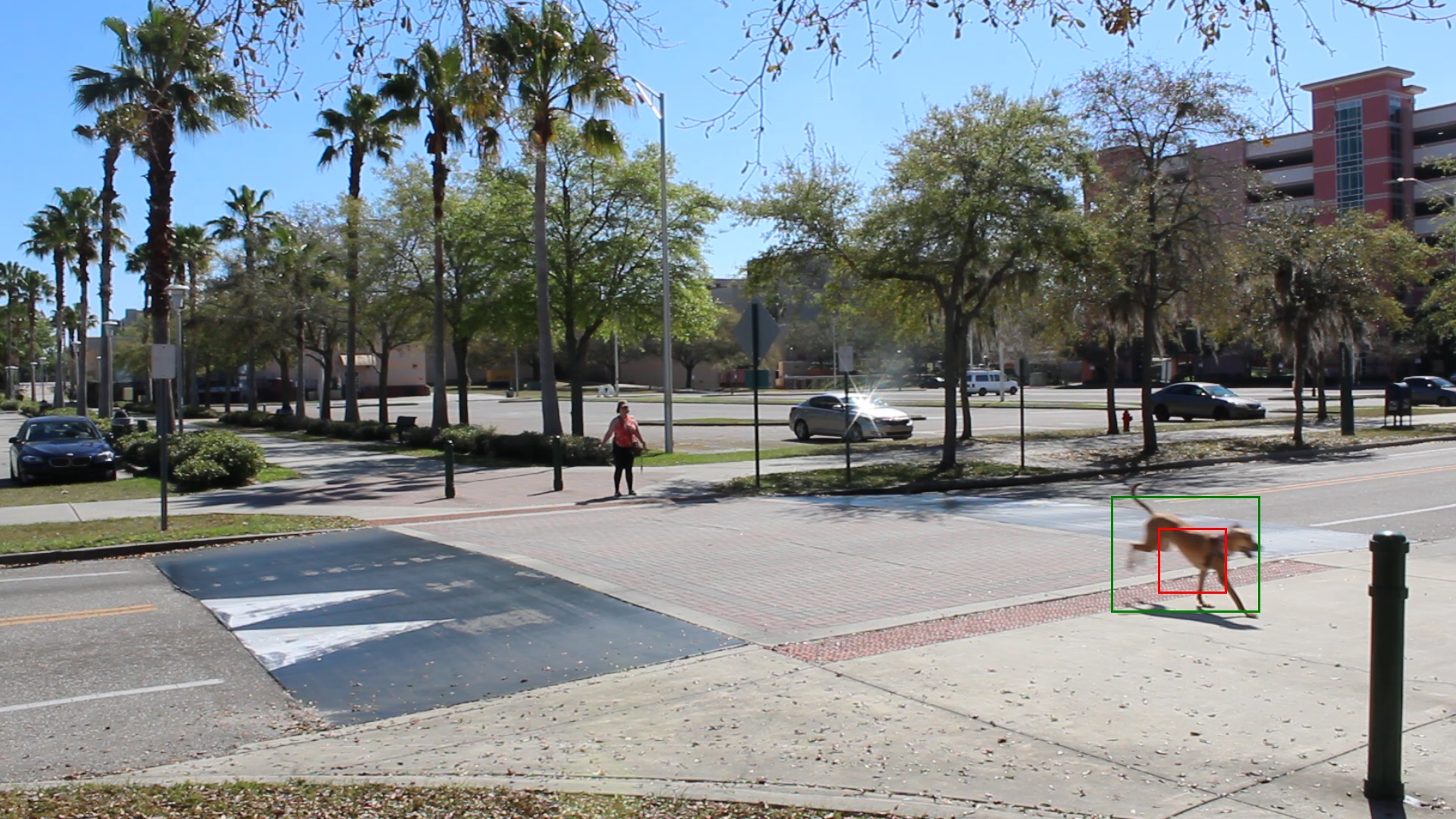}
\endminipage\hfill

\bigskip

\minipage{0.25\textwidth}
  \includegraphics[width=\linewidth]{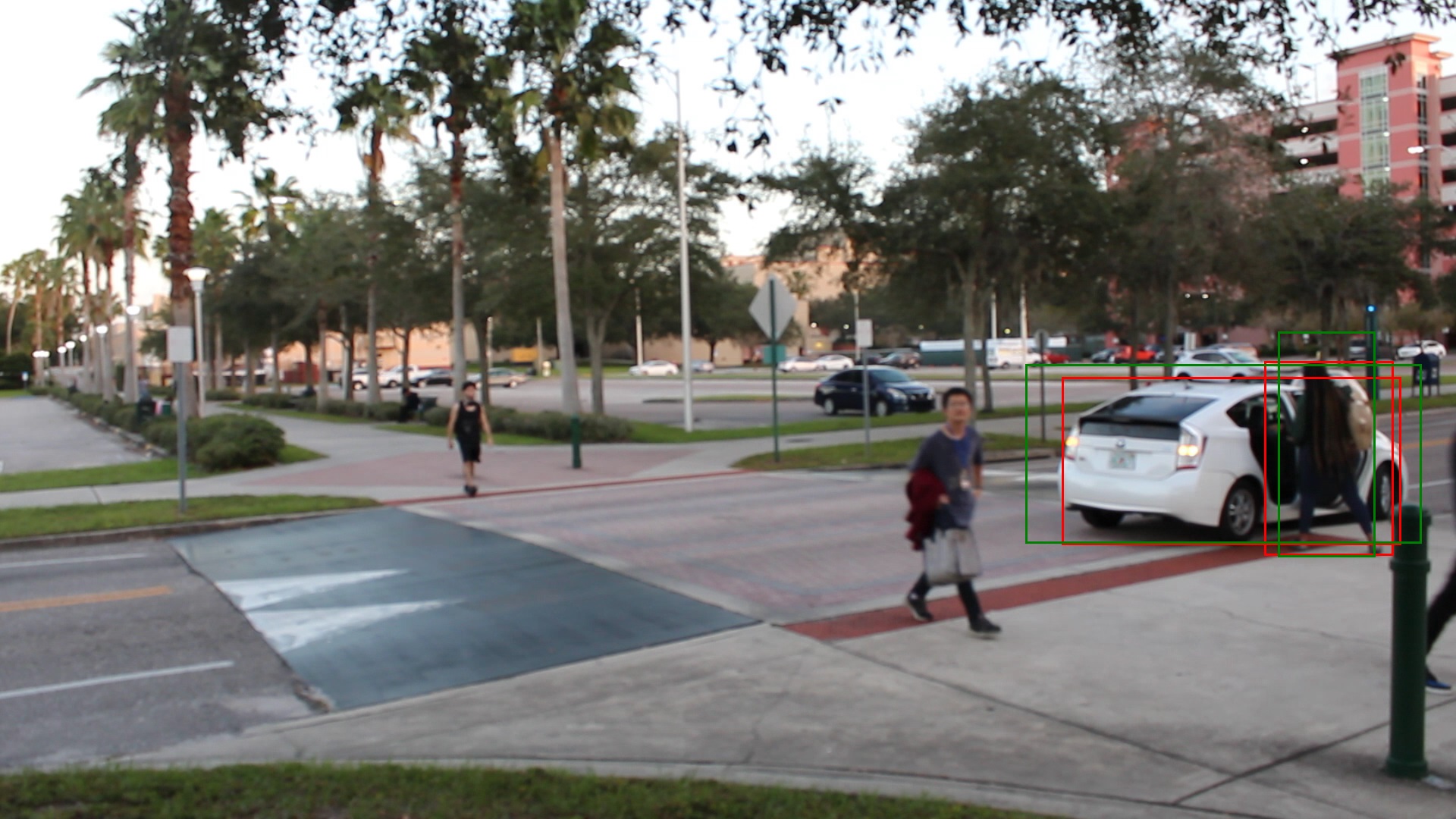}
\endminipage\hfill

\bigskip

\minipage{0.25\textwidth}
  \includegraphics[width=\linewidth]{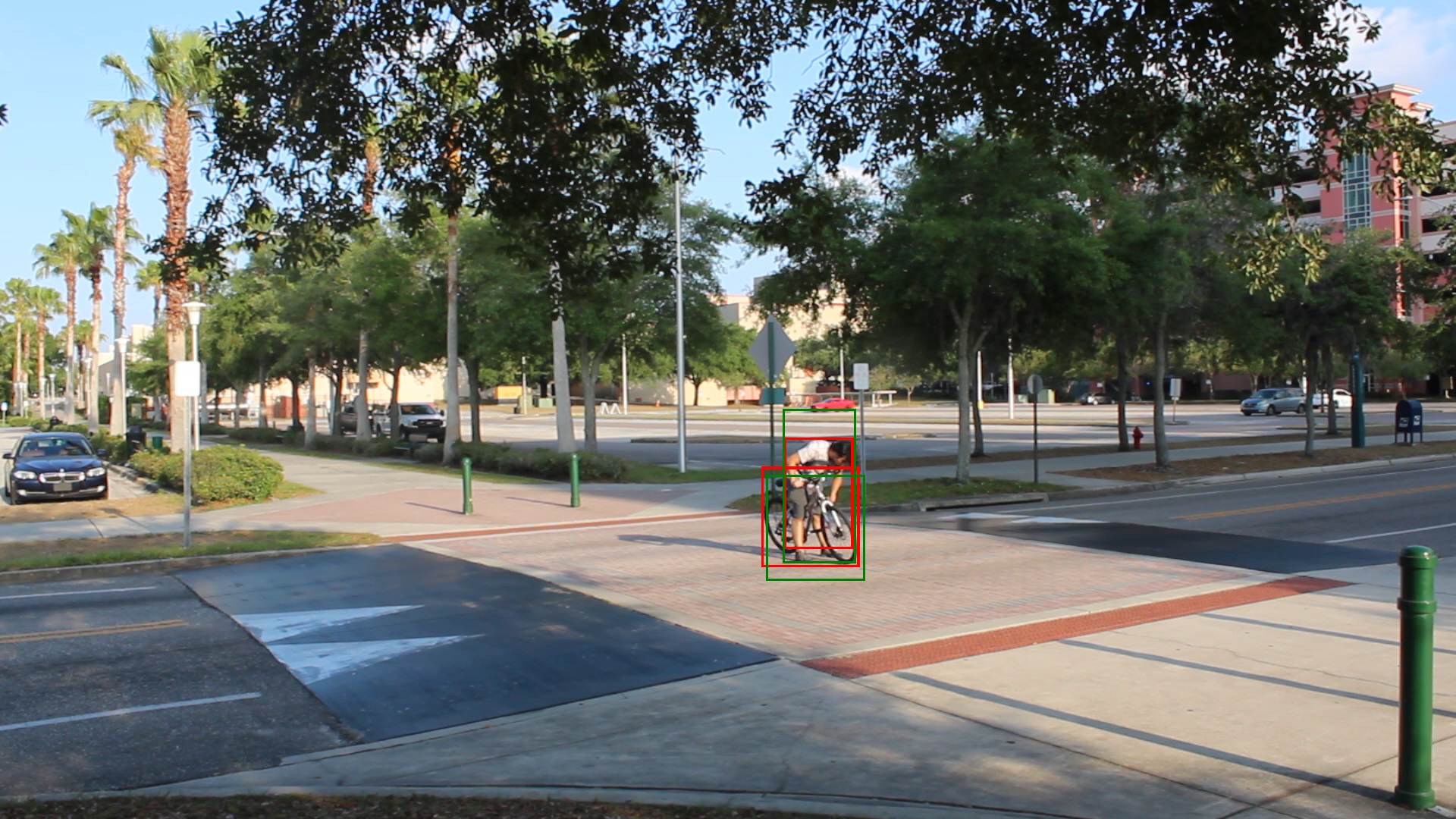}
\endminipage\hfill

\caption{Detected interaction anomalies with our method. (Top) A dog without a walker. (Middle) Car picks up a passenger on crosswalk. (Bottom) Bicycle stops briefly in the middle of the road. Ground truth labels and detection boxes are represented with green and red colors, respectively.}
\label{fig:anomdets}
\vspace{-5mm}
\end{figure}

The main results of our method as well as the EVAL \cite{SinghEtAl2023} and MemAE \cite{MemAE2019} methods using the three different evaluation criteria described above are reported in Table \ref{tab:comparison-auc}.  We can see that our scene-graph based method outperforms the other two recent methods under all criteria.  The MemAE method does very poorly for the two criteria that measure spatial localization.  This implies that the regions of an image that MemAE predicts as anomalous are usually normal.

We also show some visualizations of the output of our method on some frames from ComplexVAD in Figures \ref{fig:droppedbag} and \ref{fig:anomdets}.  Figure \ref{fig:droppedbag} shows 5 frames from a test video in ComplexVAD in which a person carrying an object places the object on the ground and continues walking.  This is an example of a "left-behind object" anomaly and is correctly detected by our method.  Figure \ref{fig:anomdets} shows frames from three other anomalies, including a dog walking without a person holding its leash, a person getting into a car in the middle of a crosswalk, and a person stopping on a bicycle in the middle of a crosswalk.  These are all successfully detected by our method.  The first anomaly is particularly interesting because it required the system to notice that in the nominal training videos, dogs always appeared with a person walking them on a leash.  It is the lack of the expected interaction that is anomalous here.

\section{Future Work and Discussions }
Complex video anomaly detection is a new direction in research and according to the baseline results, there is plenty of room for improvement for this difficult problem. Since the limitations of the object detector directly affect our method's accuracy, investigating the effects of different object detectors may lead to improved accuracy.  Also, because our method only models the interactions of pairs of objects, expanding this to modeling three or more objects interacting may also lead to accuracy gains. Another interesting direction for further research is explainability.  As shown by other papers \cite{SinghEtAl2023,ChenEtAl2018,doshi2023interpretable,SunEtAl2020}, the use of object-level models and scene graphs allow for human-understandable explanations to be automatically generated to explain why certain activities are detected as anomalous.

Our interest in introducing complex video anomaly detection is to make this research area more applicable in the real world.  An important practical issue that real systems must handle is adversarial attacks which have been demonstrated to effectively deceive video anomaly detection systems \cite{mumcu2022adversarial}. Therefore, robustness against such attacks should be a major concern in this new field.

\section{Conclusion}

Existing video anomaly detection datasets demonstrate anomalous activities that mainly involve a single object or actor. However, in the real world, anomalies are often caused by the interactions between objects. In this work, we introduce a new video anomaly detection dataset, ComplexVAD, with many diverse types of interaction-based anomalies. With the introduction of ComplexVAD, we anticipate that more research will be directed towards detecting complex anomalies in video. 

In addition to a new dataset, we also introduce a novel method to detect complex anomalies. With our method and two other state-of-the art video anomaly detection methods, we provide baseline scores on ComplexVAD. Results indicate that our method outperforms the existing methods but there is still room for improvement with further research.  

\section*{Acknowledgement}
We sincerely thank Lokman Bekit \textit{(lbekit@usf.edu)} for his valuable contributions to the data collection process.

{\small
\bibliographystyle{ieee_fullname}
\bibliography{egbib}
}


\setcounter{equation}{0}
\setcounter{section}{0}
\setcounter{figure}{0}
\setcounter{table}{0}
\makeatletter
\renewcommand{\theequation}{S\arabic{equation}}
\renewcommand{\thefigure}{S\arabic{figure}}

\clearpage
\maketitlesupplementary

\section{More details on ComplexVAD}

\subsection{Dataset Details}

In this section, we provide additional details about the dataset.

{\bf Motivation} The ComplexVAD dataset was created to encourage new solutions to the video anomaly detection problem, and in particular, to encourage methods that can handle anomalous interactions among people and objects which often occur in real-world scenarios.  The collection and labeling of the dataset was done as a collaboration by researchers at the University of South Florida and Mitsubishi Electric Research Laboratories and funded by the University of South Florida and Mitsubishi Electric Research Laboratories.

{\bf Composition} The dataset is comprised of three directories.  The "Train" directory contains 104 MPEG videos of a single scene taken in a public space on the campus of the University of South Florida (USF).  The videos in the Train directory define normal activity for this scene.  The scene shows a two-lane street with a pedestrian crosswalk going across it as well as sidewalks on either side of the street.  Car parking lots are also visible in the background.  The "Test" directory contains 113 MPEG videos of the same scene on the USF campus.  Videos in the "Test" directory contain one or more anomalous activities such as a person leaving behind a package, a cyclist colliding with a pedestrian or a person sitting on the hood of a car.  The "annotations" directory contains 113 JSON files (one for each test video) with ground truth annotations for all anomalies in each test video.  The format of an annotation file is as follows:

\{ \\
"total\_frame": ...,\\
"annotations": [ \\
\hspace*{0.2in}	\{ \\
\hspace*{0.2in}	"track\_id": ...,\\
\hspace*{0.2in}	"frame\_id": ...,\\
\hspace*{0.2in}	"bbox": ...,\\
\hspace*{0.2in}	"object\_type": ...\\
\hspace*{0.2in}	\},\\
\hspace*{0.2in}	\{\\
\hspace*{0.2in} "track\_id": ...,\\
\hspace*{0.2in}	"frame\_id": ...,\\
\hspace*{0.2in}	"bbox": ...,\\
\hspace*{0.2in}	"object\_type": ...\\
\hspace*{0.2in}	\},\\
\hspace*{0.2in}	...]\\
\}\\
where total\_frame represents the total number of frames in the video. The annotations field contains the list of each annotated object in every frame with the following properties:

track\_id: unique id for the object\\
frame\_id: frame number of the object\\
bbox: bounding box of the object in the format of [x1, y1, x2, y2] where (x1, y1) is the coordinate of top-left and (x2, y2) is the coordinate of top-right for the bounding box\\
object\_type: type of the object i.e., person, skateboard.

Note that a unique track\_id represents the same object through different frames. If a particular object is present in consecutive frames, the corresponding annotations will have the same track\_id with different frame\_id and bbox values.

Videos were collected at various times during the day and on each day of the week.  Videos vary in duration with most being about 12 minutes long.  The total duration of all training and testing videos is a little over 34 hours.  Each frame has a resolution of 1920 pixels wide by 1080 pixels high.

The videos in the Train directory should be used to learn a model of normal activity for the scene.  Videos in the Test directory should be used for trying to detect anomalous activity (activity that does not occur in any training video).  The annotations are used for evaluating the accuracy of anomaly detection using the region-based detection criterion \cite{ramachandra2020street}, track-based detection criterion \cite{ramachandra2020street} or frame-level criterion \cite{mahadevan2010anomaly}.

{\bf Collection Process} All videos were collected using a Canon EOS Rebel T6 video camera set on a tripod on the USF campus.  Videos are stored as MPEG files using an MPEG-H Part 2/HEVC (H.265) (hev1) codec. 
 Frame resolution is 1920x1080 pixels and videos are recorded at 30 frames/second.
 Videos were collected over many different days over a 5 month period in 2023.  On each day that video was collected, the camera was positioned in approximately the same way so that approximately the same area was in view for every video.  For nominal videos in the Train directory, the camera simply recorded naturally occurring activities in the scene.  For videos in the Test directory, some videos were acquired from naturally occurring activity that happened to capture unusual events while others were acquired by actors who purposely created anomalous interactions.

The Institutional Review Board at USF was consulted about the collection of video in a public space and concluded that because the "project does not include interacting with the individuals in the recordings to collect information, then it does not meet the definition of Human Subjects Research and does not require submission of an application for the IRB’s review."

{\bf Preprocessing/cleaning/labeling}
In order to preserve the privacy of people captured in the videos, a face detector \cite{li2019dsfd} was run on every frame and any detected faces were blurred with a Gaussian kernel.

The annotations for all anomalies in the Test videos consist of bounding boxes around each person/object involved in the anomalous activity as detailed above.  The annotations were manually created using the Computer Vision Annotation Tool (CVAT) (https://www.cvat.ai).

{\bf Distribution} The ComplexVAD dataset can be freely downloaded from:

https://www.merl.com/research/downloads/ComplexVAD.
It is distributed under the CC-BY-SA-4.0 license.

{\bf Maintenance} ComplexVAD is maintained by Mike Jones at MERL who can be contacted regarding any questions about the dataset. 

\subsection{Anomaly Types}

\begin{figure}
  \centering
   \includegraphics[width=\linewidth]{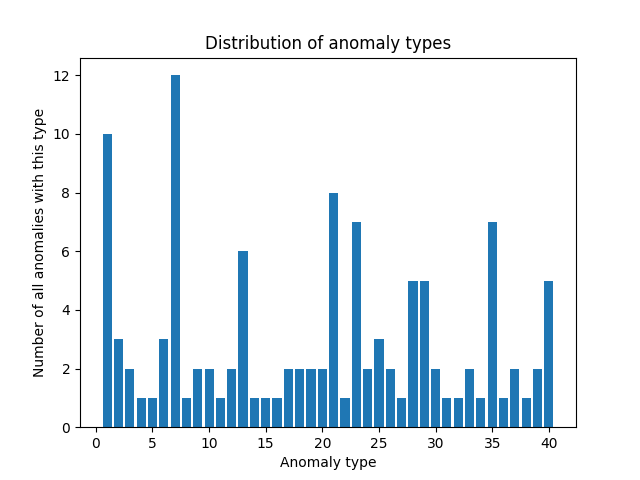}
   \caption{Numbers of each anomaly type represented in the ComplexVAD.  The numbers along the x-axis are the anomaly type indices listed in the paper.  The top three most common anomaly types are: 7- Person falling from a skateboard, 1 - Person leaving an object on the ground, and 21 - Skateboarder uses the main road.}
   \label{fig:numbers}
\end{figure}

The ComplexVAD dataset includes many different types of anomalies, many of which involve interactions among two objects or actors. Figure \ref{fig:numbers} demonstrates the numbers of each anomaly type represented in the ComplexVAD for the following list of anomaly types represented in the dataset:

\begin{enumerate}
    \item Person leaving an object on the ground
    \item Person blocking a car
    \item Car hitting a person
    \item Bicycle hitting a person
    \item Person trying to break into a car
    \item Person sitting on a car
    \item Person falling from a skateboard
    \item Piggyback
    \item Dog not on leash
    \item Two people fighting
    \item Person pushing someone to the road
    \item Runner colliding with another person
    \item Car breaking hard to stop for pedestrian
    \item After stopping for pedestrian car unexpectedly moves, which makes the pedestrian run
    \item Pedestrian preparing to cross the street has to stop because car does not stop
    \item Imitating vandalizing a car (e.g., with a long stick or baseball bat)
    \item Person hitting a tree with baseball bat
    \item Person nailing something to a tree
    \item Multiple people suddenly running scattered around
    \item Skateboard moves on its own without a user
    \item Skateboarder uses the main road
    \item Person hits someone with a bat, takes his 
    wallet, then runs
    \item People play with a ball in the middle of the street
    \item Students kick soccer ball across the street
    \item Two men carry bike
    \item Two bikers bump each other
    \item Two skateboarders bump each other
    \item Two people ride one scooter
    \item Scooters, bikes left alone
    \item Person leaves an object on top of car
    \item Person walking with unusual path
    \item Man tries to climb a pole
    \item A golf cart with a trailer stops and waits 
    \item Woman pushing a trolley
    \item Person pushes a skateboard with his feet while skateboard has bag on it
    \item Person carries another person with a trolley
    \item Person ties shoelace in the middle of the street
    \item Person falling while running/walking
    \item Biker going on a non-straight path (e.g., taking a u-turn)
    \item Skateboarder going on a non-straight path (e.g. u-turn)
\end{enumerate}

\subsection{Distribution of objects in ComplexVAD dataset}

\begin{figure*}[t]
  \centering
  \minipage{.8\textwidth}
   \centering
   \includegraphics[width=\linewidth]{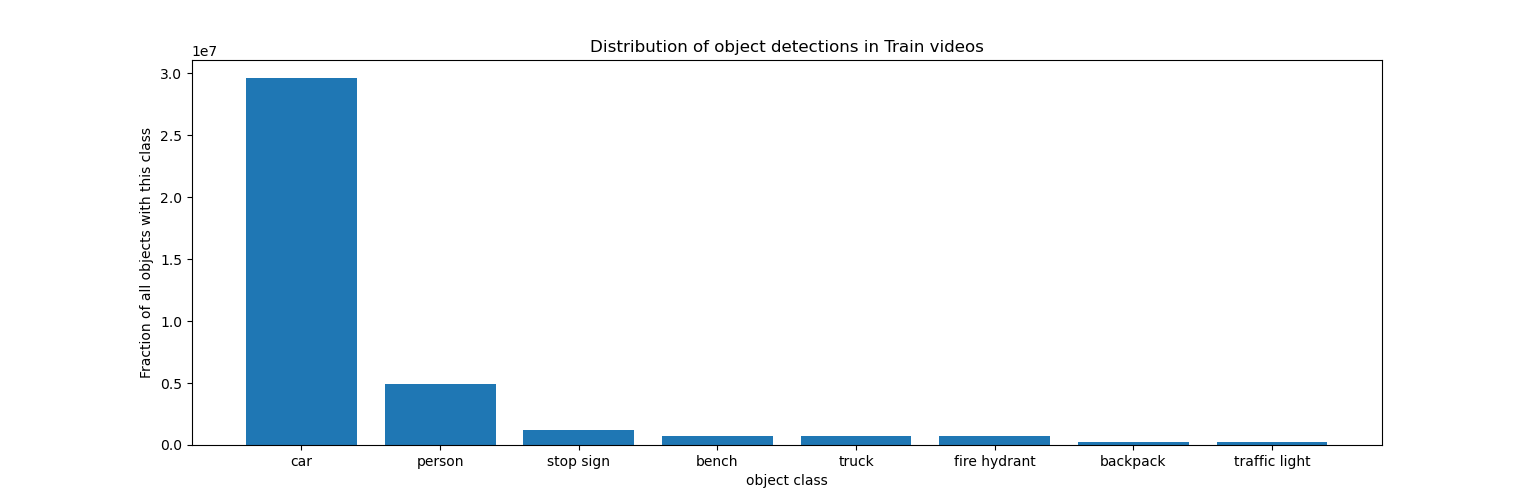}
   \endminipage \hfill
   
  \centering
  \minipage{0.8\textwidth}
   \centering
   \includegraphics[width=\linewidth]{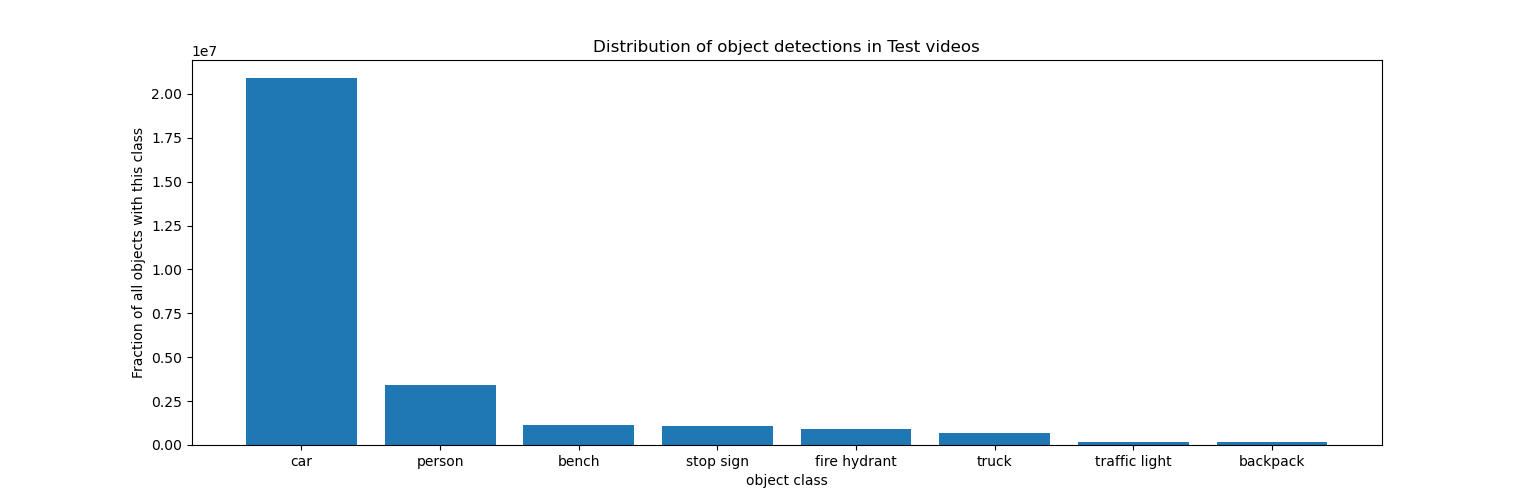}
   \endminipage \hfill
   
   \caption{Numbers of each object class detected in the Train (Top) and Test (Bottom) videos for the top eight classes.}
   \label{fig:objectdist}
\end{figure*}

To give some more insight into the contents of the ComplexVAD dataset, Figure \ref{fig:objectdist} shows bar graphs of the number of detections for the top 8 object classes detected in the Train and Test videos.  The Detectron2~\cite{wu2019detectron2} object detector which was trained on the 80 classes from MS-COCO~\cite{lin2014microsoft} was used to detect objects in each frame of the Train and Test videos.  There were a total of 38,754,900 objects detected in the Train videos and 28,847,159 objects detected in the Test videos.  Cars are the most common object detected due to the parking lot in the background of the scene and people are the second most common object class.

\section{Further details on model building}

{\bf Normalization Constants}\\

The five attribute distances in Equation (8) need to have similar scales so that one does not dominate the others.  To insure this, each attribute distance is normalized by subtracting the mean and dividing by the standard deviation.  We use pairs of nodes computed from the nominal video of a dataset to compute each attribute's distance distribution for that dataset.  The resulting normalized distances are less than 0 if two nodes are very similar (raw attribute distance less than the mean), and greater than 1 if two nodes are significantly different (raw attribute distance greater than the mean plus standard deviation).

\vspace{0.1in}
\noindent
{\bf Selecting exemplars across all nominal videos}\\

The model building process described so far selects sets of exemplars (for isolated nodes and node pairs) for a single nominal video.  Because most datasets, including ComplexVAD, include multiple nominal videos, we need a way of selecting exemplars across all nominal videos.  To do this, we simply take the union of all the exemplar sets selected for each nominal video (again, independently for isolated nodes and node pairs).  Then we run exemplar selection again over the union set.  This effectively removes similar exemplars in the union set and leaves a final set of exemplars that cover the variety of exemplars found in all nominal videos.  The final result is a set of isolated node exemplars denoted $\mathcal{E}_{iso}$ and a separate set of node pair exemplars denoted $\mathcal{E}_{pair}$ across all nominal videos.

\section{Visualizations of results}

We have included visualizations of the anomaly detections for our new method as well as the EVAL \cite{SinghEtAl2023} and MemAE \cite{MemAE2019} methods on subsequences from 6 different test videos from the ComplexVAD dataset.  Each subsequence contains an anomalous event. 
\footnote{\url{https://merl.com/research/highlights/ComplexVAD/2_supp.zip}}

The result videos for our method (filenames beginning with "Ours") and for the MemAE method (filenames beginning with "MemAE") show green bounding boxes around annotated ground truth anomalies and red bounding boxes around detected anomalies.  The MemAE result videos are much lower resolution and are grayscale because this is the input to the MemAE algorithm.  The result videos for the EVAL method show regions detected as anomalous shaded in red.  Ground truth annotations are not visualized in the EVAL result videos.

We will discuss each result video individually below, but we first make some general comments.  The result videos show that our method generally does a good job in detecting anomalous activity and has relatively few false positive detections.  In some cases, our method detects anomalous activity that is not marked in the ground truth annotations but can reasonably be regarded as anomalous.  For example, a person loitering around the cross-walk when there are no cars coming.  This did not occur in the nominal videos, but was not marked as anomalous in the ground truth annotations. Manually annotating anomalies is difficult due to many ambiguous cases.  For the EVAL method, it detects many of the anomalies, but also has many more false positives than our method.  Furthermore, its localization of anomalies is much looser than ours due to EVAL's use of a grid of fixed-sized regions instead of the object detections that we use.  For the MemAE method, it does a poor job of localizing anomalies and also has very many false positives especially in the tree branches for which there is a lot of movement due to wind.

In the following, we discuss each result video individually.

\noindent
{\bf video 4344:}
This video shows a person crossing the street at the cross-walk and then suddenly kneeling down in the middle of the street.  The person then gets back up and continues walking.  Our method does a good job of detecting this anomalous activity both temporally and spatially with no false positives.  The EVAL method also detects well with no false positives although its detections are much looser around the person.  The MemAE method fails to detect the anomaly and has many false positive detections in the swaying tree branches.

\noindent
{\bf video 4371:}
This video shows a person on a bike and a person on a scooter (slowly) bump into each other in the middle of the cross-walk and then go around each other to continue moving across the street.  Our method detects both pairs of objects (person-bike and person-scooter) for much of the anomalous interaction.  It only has a few, small false positive detections on people walking at the right of the frame near the end.  The EVAL method also detects the anomalous activity, but has quite a few false positive detections in the trees and other areas as well as continuing to detect the person and bike while they are moving normally after the anomalous interaction.  The MemAE method fails to detect the anomalous event and has many small false positives especially in the swaying tree branches.

\noindent
{\bf video 4376:}
This video shows a person loitering on the sidewalk in front of the cross-walk and then a person riding a bike nearly runs into him.  The person moves out of the way and the biker continues across the street.  Our method detects a good proportion of the anomalous activity as anomalous.  It also detects the person loitering as anomalous.  Even though this is not marked as a ground truth anomaly, it can be considered anomalous because it does not occur in the nominal videos.  Our method has a few small false positives on a car in the background.  The EVAL method also does a good job of detecting the anomalous interaction and also detects some instances of the person loitering, but it has many more false positives than our method.  The MemAE method again fails to detect the anomaly and has many false positives.

\noindent
{\bf video 4379:}
This video shows a biker riding across the street in the cross-walk, but then unexpectedly stopping in the middle of the cross-walk before continuing across the street.  Our method does a good job of correctly detecting the stopped biker with no false positives.  The EVAL method also detects the anomaly well, but has a few false positives.  The MemAE method has a few very small detections on the stopped biker but does a poor job of spatially localizing this anomaly.  It also continues to have many false positive detections.

\noindent
{\bf video 4383:}
This video shows a person walking his bike across the street and then stopping and parking the bike on the sidewalk and then walking away from the bike.  The ground truth annotation marks the person stopping and parking his bike as anomalous as well as marking the left-behind bike as anomalous.  Our method detects some of the instances of the person stopping and parking his bike as anomalous and also detects the left-behind bike as anomalous.  There are a couple of short-lived false positive detections.  The EVAL method fails to detect any of the anomalous activity (parking the bike on the sidewalk or leaving the bike behind) and has a larger number of false positives.  The MemAE method does not detect the anomalous activity and has many false positives.

\noindent
{\bf video 4398:}
This video shows a person loitering on the sidewalk with a soccer ball.  Then a skateboarder comes and runs into the soccer ball followed by the skateboarder, the soccer ball and the person all crossing the street.  Here once again, what to annotate as anomalous is ambiguous.  Only the skateboarder running into the soccer ball is marked as anomalous.  However, the person loitering with the soccer ball and the skateboarder and soccer ball crossing the street near each other could also be considered anomalous.  Our method correctly detects much of the annotated anomaly but also detects the person and the soccer ball that are stationary at the beginning of the video as anomalous.  It also detects the skateboarder and soccer ball traveling together across the street as anomalous.
The EVAL method fails to detect most of the skateboarder running into the soccer ball.  It does detect the person loitering at the beginning as well as some of the person and soccer ball crossing the street which is arguably anomalous.  EVAL also has a few more false positives than our method.  The MemAE method once again does not detect the anomalous activity and continues to have many small false positives all around the image and especially in the swaying trees.

\section{Visualizations of object attributes and closest matching exemplar}

\begin{figure}[t]
  \centering
   \includegraphics[width=\linewidth]{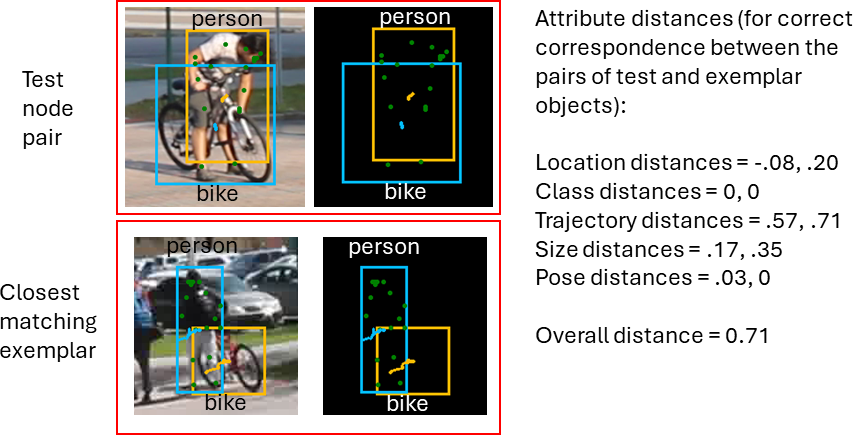}
   \caption{Visualization of a test node pair with four of its attributes: class ID, trajectory, size, pose (the location within the frame is not visualized here).  }
   \label{fig:visualization}
   
\end{figure}

Figure \ref{fig:visualization} shows in the top, left a pair of interacting objects (person and bike) from a test video that are detected by our method and linked due to proximity.  This test node pair represents an anomalous person and bike that have stopped in the middle of the road.  Below the test node pair is a visualization of the closest matching exemplar node pair.  The closest matching exemplar pair is also a person and bike but the person is walking the bike toward the left of the frame.  For each object, the size is indicated by the bounding box and the class ID is written above or below the box.  The trajectory for each object is visualized by a sequence of 30 dots (one for each of the 30 frames that it is tracked) of the same color as the object's bounding box and starting from the middle of the box.  For these test objects which are barely moving, the trajectory is very short.  The 17 coordinates that comprise the pose of a person are shown as green dots.  The same set of attributes are shown on the left of each visualization overlayed on the original frame and then again on the right over a black background so that they are more easily seen.
The figure also shows the attribute distances between the test node pair and the closest matching exemplar pair for the correct correspondence between objects in the pairs.  From this we can see that the trajectory distance (0.71) is the largest and is assigned as the anomaly score according to Equation 8 in the main paper.  This distance is above the anomaly threshold of 0.5 which results in the test person and bike being detected as anomalous.  This means that there was no pair of person and bike found in the nominal training videos with a similar trajectory (stationary).  This is because all people and bike pairs in the nominal videos were moving across the road and not stopped in the middle of the road.  We can easily use this information to provide a simple explanation of why this pair of person and bike was detected as anomalous.

\section{More insights about our method's performance on ComplexVAD}

\begin{figure}
  \centering
   \includegraphics[width=\linewidth]{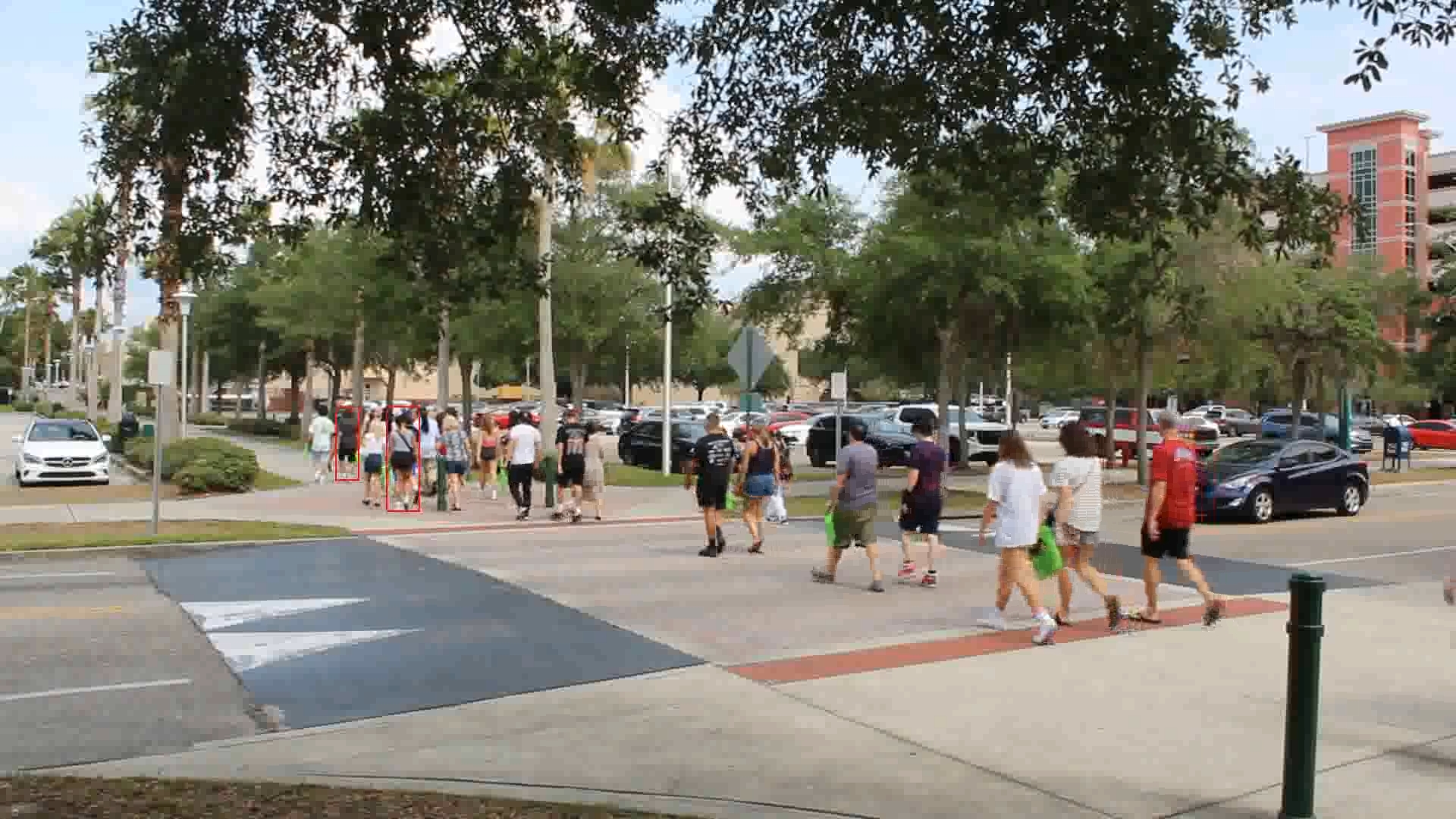}
   \caption{Crowded scene example. Red bounding boxes show the false positives our method raises momentarily.}
   \label{fig:crowd}
\end{figure}

\begin{figure}
  \centering
   \includegraphics[width=\linewidth]{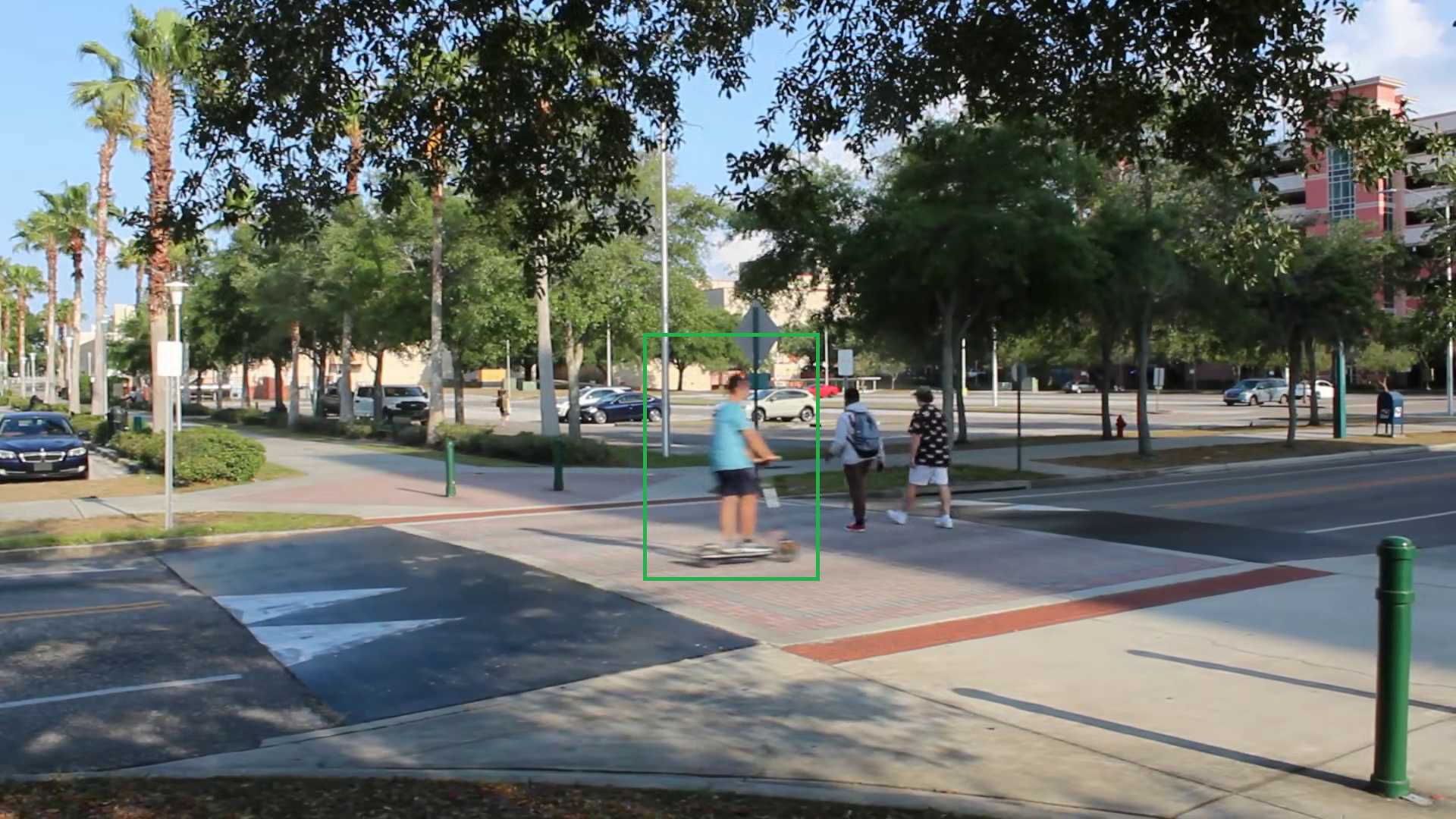}
   \caption{Simple anomaly example. Skateboarder goes on the road. Similar location-based simple anomalies are commonly found in the existing datasets.}
   \label{fig:simple}
\end{figure}

\begin{figure}[h]
\centering
  \minipage{0.5\textwidth}
   \centering
   \includegraphics[width=\linewidth]{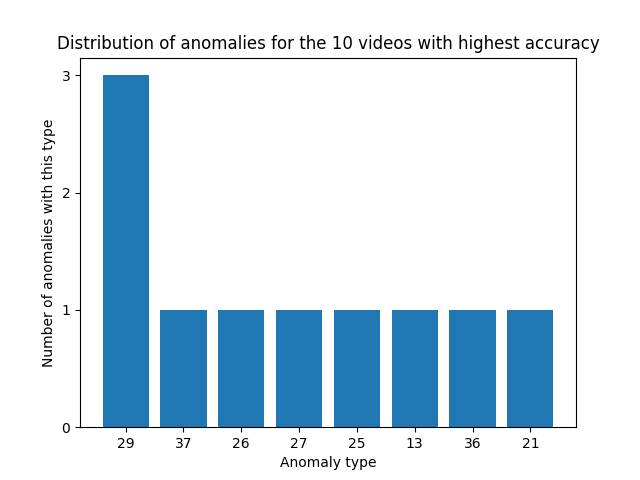}
   \subcaption{}
   \endminipage \hfill
   \minipage{0.5\textwidth}
   \centering
   \includegraphics[width=\linewidth]{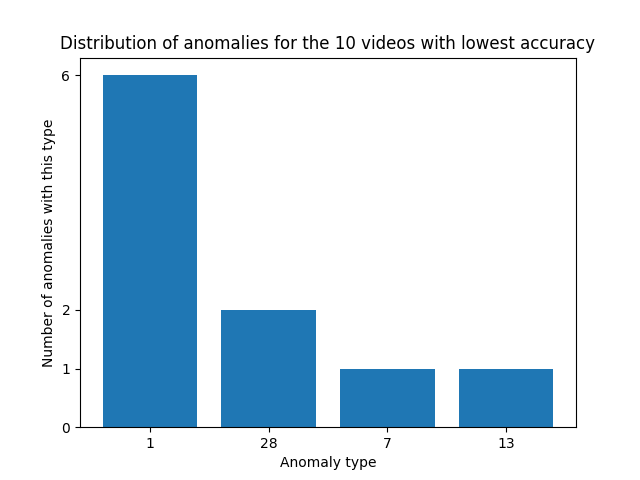}
   \subcaption{}
   \endminipage \hfill
   
   \caption{Distribution of anomalies for the 10 videos with highest (a) and lowest (b) accuracy for the Scene-Graph method.  The most common anomaly type in the highest-accuracy videos was 29 - Scooters, bikes left alone.  The most common anomaly type in the lowest-accuracy videos was 1 - Person leaving an object on the ground.}
   \label{fig:anomalydist}
\end{figure}

Figure \ref{fig:anomalydist} shows the distribution of anomalies for the 10 videos with highest (a) and lowest (b) accuracy for our method. According to the results, the most common anomaly type in the highest-accuracy videos was 29 - Scooters, bikes left alone while the most common anomaly type in the lowest-accuracy videos was 1 - Person leaving an object on the ground.

Crowded scenes is one of the challenging aspect of ComplexVAD dataset and may cause difficulties for object based methods which detect and track objects. Figure \ref{fig:crowd} shows false positives raised by our method.

In addition to interaction based complex anomalies, ComplexVAD also includes non-interaction-based simple anomalies, such as the example shown in Figure \ref{fig:simple}. In this specific example, the skateboarder goes on the road, which is a simple non-interaction anomaly, similar to the location-based anomalies commonly found in the existing datasets. The Scene Graph method addresses simple anomalies by analyzing single objects. Equation (10) is the distance function, which is primarily designed to detect simple anomalies.

\end{document}